\documentclass[times, biblatex, linguex]{sp}

\let\bm\boldsymbol 
\usepackage[normalem]{ulem}
\usepackage{multicol}
\usepackage{xurl}

\usepackage{tabto}
\NumTabs{10}
\newcommand{\mytab}[1][\TabPrevPos]{\tabto{#1}\tabto{\CurrentLineWidth}}
\newcommand{\mytabhere}[0]{\tabto{\CurrentLineWidth}}

\newcommand{\tp}[1]{\ensuremath{\left\langle #1 \right\rangle}}

\newcommand{\ml}[1]{\text{\sc #1}}

\newcommand{\svs}[2][g]{\ensuremath{{\left\llbracket #2 \right\rrbracket^{#1}}}}
\renewcommand{\sv}[1]{\svs[]{#1}}

\newcommand{\rv}[1]{{^{X}\!}}

\newsavebox{\ulcornerbox}
\begin{lrbox}{\ulcornerbox}
\raisebox{1ex}{\scalebox{0.6}[0.6]{\rotatebox[origin=c]{90}{\(\neg\)}}}
\end{lrbox}
\newsavebox{\urcornerbox}
\begin{lrbox}{\urcornerbox}
\reflectbox{\usebox{\ulcornerbox}}
\end{lrbox}
\newcommand{\qv}[1]{\ensuremath{\usebox{\ulcornerbox}#1\usebox{\urcornerbox}}}

\title[Intensional anaphora]{Intensional anaphora%
  \thanks{Authors are listed in reverse alphabetic order. Both authors
    contributed equally to all aspects of the research and preparation
  of the work.
  We thank Richmond Thomason for comments and for fielding our
  questions on logical points. 
  Ezra Keshet delivered a talk based on this material
  at the University of Texas at Austin, and we thank the audience for
  questions and comments. All remaining errors and infelicities are of
  course our own.
}
}

\author[Keshet, Abney]{
  \spauthor{Ezra Keshet \\ \institute{University of Michigan}} \AND
  \spauthor{Steven Abney \\ \institute{University of Michigan}}
}

\spyear{2024}
\spvolume{17}
\sparticle{9}
\splastpage{54}

\hypersetup{%
pdfauthor={Ezra Keshet and Steven Abney},
pdftitle={Intensional anaphora},
pdfkeywords={improper anaphora, anaphora in intensional contexts,
  modal subordination, donkey anaphora, dynamic semantics}}
\copyrightowner{Ezra Keshet and Steven Abney}

\begin{document}
\makeatletter\thispagestyle{spfirstheadings}\makeatother

\maketitle

\begin{mshistory}
  Submitted 2023-11-01 \(/\) First decision 2024-05-05 \(/\) Revision received 2024-06-13 \(/\) Accepted 2024-06-19 \(/\) Final version received 2024-07-08 \(/\) Published 2024-07-14 \(/\) Final typesetting 2024-11-24
\end{mshistory}

\begin{abstract}
Intensional operators are often treated as quantifiers over possible worlds, parallel to the treatment of determiners as quantifiers over individuals. Individuals introduced in intensional contexts cannot serve as antecedents to later pronouns as easily as those introduced in (merely) quantificational contexts, though. For instance, a quantified sentence like \textit{Everyone is eating a cheeseburger} may be felicitously followed by an anaphoric statement like \textit{They are large}, where \textit{they} refers to the totality of cheeseburgers being eaten. However, as \citet{stone1999reference} points out, the quite similar \textit{Andrea might be eating a cheeseburger} does not support later anaphoric references such as \textit{It is large} or \textit{They are large}. \citet{stone1999reference}, \citet{stone1999dynamic}, and \citet{brasoveanu2010decomposing} solve this problem by restricting the value of pronouns: in their systems, a pronoun presupposes that its referent(s) exist in the world of evaluation, ruling out anaphora from non-veridical intensional contexts. And yet, we show in this paper (i)~cases where such anaphora is disallowed even when the pronoun's referents clearly exist and (ii)~cases where such anaphora is indeed allowed, even though the pronoun's referents might not exist. We argue instead that intensional anaphora is best captured using a \textit{description-based}, rather than a \textit{value-based} account. We propose that a pronoun presupposes that its corresponding antecedent description is instantiated in each world of the context set. For instance, there must be a cheeseburger being eaten by Andrea in every candidate world of the context set in order for \textit{It is large} to be felicitous after \textit{Andrea might be eating a cheeseburger}. We implement our proposal via a new logic (building on work by \citet{keshet2018dynamic,unif}) that we name Plural Intensional Presuppositional predicate calculus (or PIP). Each PIP formula translates directly into standard first-order predicate calculus with set abstraction, providing a classical foundation for this work.


\end{abstract}


\section{Introduction}

This paper continues a project, begun in \cite{unif}, to find the
essential building blocks in any account of natural-language
anaphora. By paring down theoretical machinery, without
sacrificing coverage, we hope to obtain insight into the 
fundamental nature of this phenomenon.

Two key additions, beyond standard first-order set theory, were found to
suffice to capture a broad range of 
canonical
improper anaphora cases, including donkey anaphora
\citep{geach1962reference}, quantificational subordination
\citep{karttunen1969pronouns,sells1985restrictive}, and paycheck
pronouns \citep{karttunen1969pronouns,jacobson2000paycheck}. These additions
were (i) \textbf{formula labels}, which store previous 
formulas for later use, and (ii) \textbf{local variables}, specially
marked variables which are later bound by non-specific existential
closure operators.

In this paper, we add the capability to analyze intensions and presuppositions,
in a way intended to be as simple, standard, and general as
possible. When we do so, we find that the predictions
for anaphora in intensional contexts are not only empirically
confirmed, but also solve
certain puzzles described by \citet{stone1999reference}.
We take this as evidence in favor of the approach.

The resulting system, combining formula labels, local variables, possible worlds, and a
presupposition operator, we call Plural Intensional Presuppositional predicate calculus
(PIP). We also take this opportunity to further streamline the system:
PIP is simply the first-order Predicate
Calculus with equality and set abstraction (PC) with a small number of defined---and hence
eliminable---constructs.

\subsection{Descriptions and values}

Semantic theories of complex anaphora generally fall into one of two
camps. \textbf{De\-scription-based} accounts make crucial use of
salient descriptions within the meanings of complex pronouns. For
instance, E-type theories, following \cite{evans1977pronouns1}, treat
certain pronouns as being akin to definite descriptions, in that they pick out a
referent that satisfies some salient property. While this approach is
quite successful and widely adopted, especially in static theories of
semantics, most description-based theories suffer from certain well-known problems and
leave the key term {\it salience\/} entirely undefined
(see \citealt{elbourne2005situations} for an excellent overview and an
exposition of one E-type theory). \textbf{Value-based} accounts
instead store and retrieve individual values for discourse
referents. For instance, dynamic semantic theories
\citep{kamp1981truth, heim1982diss, groenendijk1991dpl} treat pronouns
as variables whose values are individuals, but which
do not preserve the full original descriptions that were used to establish those values.\footnote{
To be sure, the contrast between description and value is one of
degree.
%
%
As we will see below, some systems preserve varying amounts of information
from the defining description in the form of tables of values.
}

\sloppy PIP's formula labels represent a variety of description-based anaphora, and as detailed in \cite{keshet2018dynamic} and \cite{unif}, 
they allow a straightforward way to handle paycheck pronouns, which can be problematic for
purely value-based accounts. At the same time, PIP provides
rigorously-defined antecedents for pronouns, without recourse to a
vague notion of saliency as often mooted for description-based
accounts. In short, PIP combines features of both approaches.

\subsection{Outline}

We present PIP in Section~\ref{sec:simple}, including a review of its coverage 
of examples that motivated the precursor system of
\cite{unif} (Section~\ref{sec:applications}). First, though, in the remainder of this section (\ref{the_problem}),
we present the problem of intensional antecedents noted by \citet{stone1999reference}, including
some new counterexamples for Stone's analysis; and
in Section~\ref{sec:analysis}, we show that PIP handles them correctly.%
 Finally, in Section
\ref{sec:prev}, we survey previous literature: first detailing the
issues with value-based accounts, and then presenting some other recent
hybrid accounts. These latter accounts, while not actually addressing the cases described below, are in fact much closer in spirit to the approach we espouse.

\subsection{Intensional antecedents}\label{the_problem}

Since medieval times, scholars have analyzed intensional operators,
such as modals, as quantifiers over worlds, parallel to determiners,
which quantify over individuals:\footnote{As noted in
\cite{fintel2008intensional}, \citet{sep-modality-medieval} sketches
the medieval origins of intensional quantification, and
\citet{copeland2002genesis} traces the modern incarnation of this idea
to Charles Sanders Peirce.}
\ex.
	\a. Everyone is eating a cheeseburger.\\
		$\rightsquigarrow \forall x \left(\exists b \left(\textsc{burger}(b) \wedge \textsc{eating}(x,b) \right)  \right)$ \label{extburger}
	\b. Andrea might be eating a cheeseburger.\\
	$\rightsquigarrow \exists  w \left(\exists b \left(\textsc{burger}_w(b) \wedge \textsc{eating}_w(\textsc{andrea},b) \right)  \right)$ \label{intburger}

As \citet{stone1999reference} points out, though, expressions within
the scope of a modal quantifier differ in their anaphoric potential
when compared to expressions embedded beneath quantifiers over individuals. 

In particular, expressions in the scope of a quantifier over
individuals, like \textit{a cheeseburger} in \ref{extburger} may serve
as the antecedent for \textbf{summation pronouns} (\cite{berg1996some}, so named by
  \citet{unif}) in subsequent sentences. For instance,
after an assertion of \ref{extburger}, the underlined pronouns in
\Next may refer to the totality of cheeseburgers being eaten.  The
pronoun will be singular if everyone is (atypically) sharing a huge
cheeseburger, or plural if people are eating multiple
burgers:\footnote{An anonymous reviewer points out that \ref{itburger}
might be supported by a structure where \textit{a cheeseburger} simply
scopes over \textit{everyone}.
Although this structure is certainly
possible, similar structures can be blocked using scope islands:
\ex. Every student only ever used a pencil sharpener in the PRINCIPAL's office. By the end of the year, it was quite worn down and dull.

\citet{barker2022rethinking} points out that the scope of an indefinite like \textit{a pencil sharpener} here is limited by the word \textit{only}: \Last cannot mean that there is one pencil sharpener that students only used in the principal's office, and others they used wherever they wanted. It has to mean that all student pencil sharpening happened in the principal's office. And yet, singular reference to a sharpener in the principal's office is possible in a later sentence. We maintain that this anaphora is due to a summation pronoun, although many other proposals have linked similar cases to \textit{wide-scope indefinites} \citep{fodor1982indefinites}.

} 
\ex. Everyone is eating a cheeseburger. 
	\a. \underline{It} is very large! \hfill ($\approx$ \textit{The burger they are eating is large})\label{itburger}
	\b. \underline{They} are very large! \hfill ($\approx$ \textit{The burgers they are eating are large})

However it is technically implemented, a summation pronoun is always
related to a description that occurs earlier in the
discourse. Let us call this the \textbf{antecedent description}. In
\Last, the antecedent description is \textit{burger(s) people are eating.}\footnote{From
this example, one might presume that the antecedent description need
not appear literally, but only be implicit in, in the previous discourse.
However, in the example given, we take
\(\ml{burger}(b)\wedge\ml{person}(p)\wedge\ml{eats}(p,b)\) to be the
meaning of the nuclear-scope argument of {\it every\/} at
LF.  Details are given below: see especially
sections~\ref{sec:formula_labels} and \ref{sec:gq}.
Fuller discussion of the syntax is found in \cite{unif}.
}

Notice next, though, that the same summation meaning is not possible with a modal quantifier as in \ref{intburger}:
\ex. Andrea might be eating a cheeseburger. \label{intsum}
	\a. \# \underline{It} is very large! \mytab[3ex] (Intended: \textit{The burger she might be eating is/would be large})
	\b. \# \underline{They} are very large!  \mytab (Intended: \textit{Any burgers she might be eating are/would be large})\label{intsumpl}

We would like to draw specific attention to the case of the plural summation in \ref{intsumpl}, which has not been thoroughly discussed in the literature to date.\footnote{Although do see the discussion surrounding example (174b) in \cite{brasoveanu2010decomposing}.} It is particularly difficult to understand this sentence as meaning that all possible cheeseburgers Andreas is eating, across all possible worlds of the context set, are large. And this intuition is sharpened further when the predicate involves a specific location in time and space:
\ex. \#\uline{They} are on the kitchen counter.

There is clearly no reading of \Last where the various cheeseburgers Andrea might be eating---lamb burgers, imitation meat burgers, turkey burgers, etc.---are all on the kitchen counter. Summation pronouns simply do not operate over multiple worlds at once, and this basic observation must be addressed in any semantic system meant to capture summation pronouns. (Our explanation for this fact is given at the start of Section \ref{sec:analysis} below.)

\citet{stone1999reference,stone1999dynamic}, and
\citet{brasoveanu2010decomposing} all suggest that the oddness in
\ref{intsum} arises from a presupposition that a pronoun's referents must
exist in the world of evaluation. Since Andrea's cheeseburger is
merely hypothetical, it is not an appropriate target for pronominal
reference. More precisely, rather than a single world of evaluation,
we can employ
\posscitet{stalknaker1978assertion} notion of a \textit{context set},
the set of worlds that satisfy the conjunction of common-ground knowledge and the
discourse so far.  The context set models our (imperfect) knowledge of the real
world. The modal sentence in \ref{intsum} asserts that there are
{\it some\/} candidates for the real world in which Andrea is eating a
cheeseburger, but it permits (indeed, implicates) that there are other candidate worlds
in which she is not. \citet{stone1999reference} and \citet{stone1999dynamic} require that a pronoun's referent exist across the entire context set.\footnote{
\citet{brasoveanu2010decomposing} is unclear on this point, since it is difficult to refer to the entire context set in his system.}

There is a major empirical problem with this approach. Modals often target individuals that (unlike the potential burgers above) clearly do exist in all candidate worlds of evaluation; but even in these cases, a similar pattern obtains. By way of illustration, imagine that there are exactly five, well-known candidates running for mayor (and no write-ins). Under this scenario, in all epistemically accessible worlds, one of these five (real-world) candidates will win the election. Late on election night, someone who has not checked the news recently may say \ref{election}, but clearly cannot follow up with \ref{election sing} or \ref{election plur}, even if they have strong opinions on who the possible winner(s) are among the five candidates:
\ex.There might already be a winner in the mayoral election.\label{election}
	\a. \#She is a woman. \hfill (Intended: \textit{The possible winner is a woman})\label{election sing}
	\b. \#They are women. \hfill (Intended: \textit{Any possible winners are women})\label{election plur}

For instance, the speaker may believe that there is only one viable
candidate, say Jackson, who might win the election. In this case,
Jackson is the winner in all the worlds where there already is a
winner. However, even with this one real person in mind, the speaker
cannot use \ref{election sing} to assert that Jackson is a woman. And
the same is true if there are multiple possible winners: even though
all of the candidates, and therefore all of the possible winners,
exist in the real world, this alone is not enough to allow anaphoric
reference.

More examples of this type are given here:
\ex. If there's a light on at 9pm, it is possible that one of the teachers is still inside the school building. \#She is / \#He is / \#They are working late.

\ex. If you like solving puzzles, there may be a book for you in the ``Puzzles'' bookcase.
\#It is / \#They are on the top shelf, where we keep the more serious puzzle books.

These are cases where the indefinite most naturally scopes below the modal element; the speaker does not have a particular teacher or book in mind. And yet, they involve real-world individuals: teachers from a particular school and books in a particular bookcase. Even so, the pronouns \textit{she}, \textit{he}, \textit{it}, and \textit{they} cannot refer to just the teacher(s) that might be in the building or just the book(s) that the listener might like.

The next example makes it even clearer that the plural \textit{they} is not referring to the whole class of actual items involved (like all the school's teachers or all the books in the ``Puzzles'' bookcase):
\ex. The killer has salt-and-pepper hair. If he used this car, he might have shed a hair. \#It is / \#They are white.

Again, \textit{it} and \textit{they} cannot refer just to the hair(s)
that might be in the car. And since the killer has both dark and light hair, the pronoun \textit{they} cannot refer to all the killer's hairs. Finally, as noted above, all the versions with plural \textit{they} are especially odd in these types of scenarios.

 These examples show a stark contrast between (i) existing in the
real world while satisfying an antecedent description in a
hypothetical world, and (ii) satisfying that
same antecedent description in the real world. Even when someone fitting an antecedent description in a
hypothetical world exists in the real world, that does not necessarily
mean that anyone actually fits the antecedent description in the real world:
for example, there can be plenty of potential winners without there being any
actual winner yet. 

The main intuition behind our proposal, then, is as follows: 
\ex. A pronoun presupposes that its antecedent description has a non-empty extension.\label{hypo}

Along with \posscitet{stone1999reference} standard assumption that presuppositions must hold across the context set,\footnote{Many others make similar assumptions. For instance, \citet{heim1983projection} assumes that a context $c$, construed as a proposition, must entail a presupposition $p$ in order for $c$ to admit a sentence presupposing $p$. And in order for $c$ to entail $p$, $p$ must be true in every world satisfying $c$. Similarly, once Heim extends contexts to be sets of world/assignment pairs $\tp{w,g}$, she assumes presuppositions, now denoting open formulas, must be true relative to each such pair in a given context $c$.} \ref{hypo} readily explains the examples
above: in each case, the pronoun's antecedent description has an empty extension in at least one of the worlds of the context set.

The hypothesis in \ref{hypo} also 
neatly explains certain cases that \citet{stone1999reference} captures, but \citet{brasoveanu2010decomposing} does not%
. In particular, an antecedent description that is satisfied across the whole context
set can indeed support the use of a later pronoun:\footnote{\citet{elliott2022partee} has independently made a similar point.}
\ex. There must be some sort of animal in the shed. \underline{It}'s making quite a racket!\label{animal}

Here, by using \textit{must}, the speaker is endorsing the view that
an animal is actually in the shed. That is, the antecedent
description \textit{animal in the shed} has an extension in each world
of the context set, and that extension is a unique (but potentially different) animal
in each candidate world.  The summation pronoun {\it it\/} occurring
in the following indicative sentence
refers to the value(s) of the antecedent description in the actual
world. Since the actual world is one of the candidates, the value of
the antecedent description exists %
and is unique.
Because it exists, the summation pronoun is felicitous, and because it
is unique, the singular form \textit{it} is appropriate.

Similar examples from corpora are given here, with the relevant antecedents and pronouns underlined:\footnote{Example \ref{chimp} was accessed via \cite{GloWbE}; the others via \cite{COCA}.}
\ex. [From \textit{House}, Season 5, Episode 9 ``Last Resort'' 27:20]
	\a.[HOUSE: ]  If there's swelling, it's a Pancoast tumor that's metastasized. Feel that. Right there. 
	\b.[JASON: ] (\textit{feeling his neck}) If it's cancer, there must be \underline{a test}. 
	\b.[HOUSE: ] You just did \underline{it}.

\ex. [From \textit{JAG}, Season 5, Episode 1 ``King of the Greenie Board'' 7:45]
	\a.[CHEGWIDDEN: ]  I know Commander Imes is trying a court-martial in Pensacola. So why does the watch bill say that she's on duty this weekend?
	\b.[TINER: ] There must be \underline{a mistake}, sir. 
	\b.[CHEGWIDDEN: ] And \underline{it}'s not the only one, Tiner. I can't seem to find Commander Mattoni's name on here anywhere.

\ex. [From ``Buffalo wolf''\footnote{\cite{reed2003buffalo}} by Robert Reed in \textit{The Magazine of Fantasy \& Science Fiction}, March 2003, p. 4]\\
Hoof prints showed him the recent past. With his sore finger, he touched the clearest print. [$\ldots$] It was a short-hair, he assumed. A big bull, since only bulls could be this large. There must be \underline{a ripe female} somewhere close, and \underline{she} was pulling him along by his stupid little balls [$\ldots$]

\ex. \label{chimp} [From \textit{The third chimpanzee: the evolution and future of the human animal}\footnote{\cite{diamond2014third}} by Jared Diamond]\\
Fortunately for us, the silence from outer space is deafening. Yes, out there are billions of galaxies with billions of stars. Out there must be \underline{some transmitters} as well, but not many, and \underline{they} do not last long. Probably there are no others in our galaxy, and surely none within hundreds of light-years of us.

%
%

\subsection{Other dynamic systems}\label{sec:related}
Before moving on, let us briefly explore the relationship between plural logic and the cases described. First, if one were to simply add possible worlds to a singular dynamic logic like Dynamic Predicate Logic \citep{groenendijk1991dpl}, a similar problem arises. For instance, in the formula in \Next, where $w$ is the actual world, the variable $x$ is indeed accessible outside of the existential quantification `$\exists w'$' meant to represent \textit{might}:
\ex. $\exists w' (\exists x\, \textsc{burger}(w',x)) \land \textsc{tasty}(w,x)$

Such a simple system would have to make a move parallel to Stone's and disallow this anaphora on other grounds.

Systems which add full generalized quantifiers to singular dynamic logic \citep[for instance][]{chierchia1992anaphora,eijck1992dynamic} treat these generalized quantifiers as tests, opaque for anaphora. Such systems could translate modals like \textit{might} and \textit{must} as a generalized quantifiers, thus disallowing any anaphora to indefinites within these modals. Traditional dynamic treatments of modals, following \cite{veltman1996defaults}  \citep[see also][]{groenendijk1996coreference,goldstein2019generalized}, likewise treat modals as tests, preventing any anaphora to items in their scope. Neither of these approaches would explain, however, why anaphora is indeed allowed out of \textit{must}, as shown in \ref{animal} above.

In addition, none of these systems handle summation pronouns, which include clear cases of anaphora to indefinites within the scope of a generalized quantifier. And none explain modal subordination, which can straightforwardly be treated in parallel to quantificational subordination in a plural logic that treats modals parallel to quantifiers \citep{brasoveanu2010decomposing}.\footnote{\citet{elliott2022partee} also discusses even more cases which must allow anaphora out of a modal sentence, such as his Partee Conjunctions.} 
But once summation pronouns and modal subordination are incorporated into a semantic system, the cases due to Stone pose a particularly thorny problem:  summation pronouns must be allowed to refer to an element inside an individual generalized quantifier, but never within a modal generalized quantifier. Thus, we view Stone's cases as a problem in particular for \textit{plural} semantics: a tension between allowing summation pronouns and modal subordination on the one hand and disallowing certain types of intensional anaphora on the other. 

\subsection{Conclusion}
Finally, let us take this opportunity to distinguish the cases presented in this section from a class noted by an anonymous reviewer, exemplified in \Next:
\ex. \a. Jerry might own a car, but they are still a hassle.
	\b. Jerry doesn't own a car---they are too much of a hassle!

Here, the pronoun \textit{they} refers not to any cars that
Jerry might own, but rather to the class (or
natural kind) of cars in general. \citet{chierchia1998kinds}, following
\cite{carlson1977diss}, notes that bare plurals referencing kinds
exhibit several behaviors which he calls \textit{scopeless}. Among
these, Chierchia points out that anaphoric reference to kinds seems
not to be subject to the usual scope boundaries. We take {\Last} to be an example of such
an effect, in which \textit{they} can pick up the kind \textit{cars}
without any direct binding. This also explains the apparent
counterexample to binding out of negation, shown in \Last[b]. The
pronoun \textit{they} is not bound by \textit{a car}---it would be
singular, not plural, if \textit{a car} were the antecedent---but
rather {\it they\/} refers to the kind
\textit{cars}. 

We note here that the hypothesis in \ref{hypo} makes explicit
reference to the antecedent description for a pronoun. To the extent
that this is correct, then, there must be some representation in the semantics
of the antecedent description. In the next section, we introduce
formula labels for that purpose: formula labels retrieve previous antecedent
descriptions for later use in the discourse. In
\ref{mostrecent}, we contrast our approach with another method that
has been proposed in very recent work for
representing antecedent descriptions in the semantics. 


%

\section{PIP} \label{sec:simple}

We introduce a logic called {\it Plural Intensional Presuppositional predicate
calculus\/} (PIP), based on the logics found in
\cite{keshet2018dynamic} and \cite{unif}, but extended to include
intensions and presuppositions.\footnote{\cite{keshet2018dynamic}
introduces formula labels that are similar to those of PIP, and \cite{unif}
introduces the version of unselective binding and summation pronouns
that are adopted in PIP.}  The guiding principle in the design of PIP
is to identify the bare minimum that can be added to standard
first-order predicate calculus with sets (PC) to capture the empirical
phenomena covered by a full, dynamic plural logic such as that of
\cite{brasoveanu2010decomposing}, plus presuppositions.

To that end, PIP is just standard first-order predicate
calculus with set abstraction and equality (PC), supplemented with a handful of abbreviations and annotations, and some assumptions about the model.  The complete list of abbreviations and annotations (collectively, ``PIP constructs'') is as follows:

\ex.\label{constructs}PIP Constructs\begin{samepage}
	\a. Unselective closure of bracketed variables: \(\exists\ldots[x]\ldots\)
	\b. Summation: \(\Sigma x\phi\)
	\b. Formula-label definition and use: \(X \equiv \phi\), $X$
	\b. World subscripts on predicates: \(P_w(x_1,\ldots,x_n)\)
	\b. Presuppositions: \(\phi|\psi\)\z.
\end{samepage}

The PIP constructs are all eliminable in the sense that
there is a straightforward translation from any
PIP expression to a predicate-calculus expression with the same truth
conditions.  In addition, PIP constructs play a role in defining {\it felicity conditions,\/}
which we take to be independent of truth conditions.
\enlargethispage{\baselineskip}
As for models, we assume that they have the standard form \((D,I)\),
but satisfy the following:
\ex.\a.\label{model-a} The members of the domain $D$ are {\it pluralities,\/} which
are sets of abstract elements that we call {\it points.\/}
{\it Worlds\/} constitute a subset of the singleton pluralities
\({\mathcal W}\subseteq D\), with \(|w|=1\) for all \(w\in {\mathcal W}\).
\b.\label{2b} There is a family of valuation functions $\{V_w\}$ indexed by
world, and \(I(P)(w,x_1,\ldots,x_n) = V_w(P)(x_1,\ldots,x_n)\).

\par
In this section, we present an informal but detailed description of PIP,
explaining each of the constructs in \ref{constructs}.

\subsection{Unselective closure and bracketed variables}
Let us begin with the treatment of cross-sentential anaphora proposed
by \citet[Ch 2]{heim1982diss}, following work by \citet{lewis1975AQ}.  At the risk of putting words into their mouths, we can
characterize this proposal as introducing an unselective existential
closure operator, and a method of slating certain variables
for closure, ideas that are incorporated into PIP.
For example, the PIP formula {\Next[a]}
translates to the standard predicate calculus as {\Next[b]}:

\ex.\label{def_exists}\a. \(\exists\,\textsc{saw}(w,[x],[y])\)\hfill[PIP]
\b. \(\exists x\exists y\,\textsc{saw}(w,x,y)\)\hfill[PC]

The brackets mark certain variables that occur free in the body of
the quantifier as {\it local\/} variables, in
contrast to the unbracketed free variables, which we call {\it external\/} variables.
(The local/external distinction exists in Lewis and Heim, though their
notation and terminology differ from ours.)
The unselective existential \(\exists\)
binds the local variables in its body but not the external variables.  (Thus
{\it semiselective existential closure\/}
would be more accurate, but we will stick with
the established terminology.)  Both the PIP expression
{\Last[a]} and the PC expression {\Last[b]} mean that there exist
individuals $x$ and $y$ such that $x$ saw $y$ in the particular world
$w$.  The variable $w$ is external and remains free; its value is supplied by context.

Heim proposes that the meaning of an indefinite noun phrase is
represented as a local variable, and the meaning of a sentence is represented as an
open formula.  For example, {\it a dog appeared\/} means {\Next[a]},
with PC translation {\Next[b]}:
\ex.\a.\label{sent} \(\textsc{dog}([d])\wedge\textsc{appeared}(d)\)\hfill[PIP]
\b. \(\textsc{dog}(d)\wedge\textsc{appeared}(d)\)\hfill[PC]

Note that a free variable is local just in case {\it any\/} free
occurrence is bracketed.  In {\Last[a]}, we might have bracketed the
second occurrence of $d$, or both occurrences, with no difference in
meaning.
Note also that the brackets have no direct effect on truth
value.  Rather, their consequences are indirect,
affecting the truth conditions of an enclosing unselective closure.

Now one has the intuition that the logical representation of
{\it a dog appeared\/} should actually include an existential closure:
\ex.\a.\label{disc} \(\exists(\textsc{dog}([d])\wedge\textsc{appeared}(d))\)\hfill[PIP]
\b. \(\exists d(\textsc{dog}(d)\wedge\textsc{appeared}(d))\)\hfill[PC]

That intuition is at odds with another one, namely, that the ``it'' in
a subsequent sentence {\it it barked\/} is just $d$.  That is, we
would like to represent the meaning of the two-sentence discourse
{\Next[a]} as {\Next[b]}:
\ex.\label{xsent}\a. A dog appeared.  It barked.
\b. \(\textsc{dog}([d])\wedge\textsc{appeared}(d)\wedge\textsc{barked}(d)\)

``It'' translates simply as $d$, and the two sentence meanings are
simply conjoined.

One way of reconciling these {\it prima facie\/} incompatible
intuitions is to move to a dynamic logic, and assume that the first
sentence introduces a new variable into the discourse context
(\citealt{kamp1981truth}, \citealt[Ch 3]{heim1982diss},
\citealt{groenendijk1991dpl}). Doing so represents a rather
drastic shift in one's logical foundation: dynamic logics are
designed for programming languages, in which formulas have not only
truth conditions but also side effects---that is, they are commands
that modify state, rather than statements that are true or false.

The idea of PIP, following \cite{unif}, is to pursue a more conservative alternative,
grounded in Heim's proposal.  Instead
of abandoning the predicate calculus, let us distinguish between the
{\it sentence\/} ``a dog appeared'' and the one-sentence
{\it discourse\/} ``a dog appeared.''  We take the meaning of
the sentence to be the open formula \ref{sent}, whereas the meaning of
the discourse is \ref{disc}, in which local variables are
existentially closed.
In the general case, we take the interpretation of a {\it discourse\/}
\(\phi_1,\ldots,\phi_n\) to be:
\ex.\label{discourse_interpretation} \(\exists(\phi_1\wedge\ldots\wedge\phi_n).\)

\noindent
This captures the same intuition that motivates the dynamic approach,
namely, that the scope of the variable is wider than the sentence,
without abandoning the simplicity of the predicate calculus.

We take the position that one's intuitions about sentence meaning in
isolation are actually intuitions about single-sentence discourses. (See \citealt{cresswell2002static} for more discussion on this issue.)
To boil it down to a slogan:

\ex. The natural unit of interpretation is the complete discourse.

\subsection{Pluralities and summation}

Natural-language noun phrases may be either singular or plural, and we
take that syntactic distinction to correspond to a semantic
distinction between singular and plural entities.
As briefly indicated in \ref{model-a}, we assume that the domain of
discourse consists of {\it pluralities\/} that are in turn
sets of {\it points.\/}  We will continue to use
the term {\it individual\/} to mean ``domain element,'' with the
consequence that {\it individual\/} and {\it plurality\/} are
synonymous.  Singular noun phrases denote singleton
individuals, and plural noun phrases denote plural individuals, or ``groups,'' which are
pluralities whose cardinality is greater than one.  There is also a null
plurality \(\varnothing\), which may occur as the meaning of a quantificational
expression, but is unsuitable as the meaning of a denotational noun
phrase because of presuppositions (Section~\ref{sec:felicity}).


Most one-place predicates that we will use are distributive in the sense
that the predicate is true of a group just in case it is
true of every singleton individual in the group.  (We will call these
the {\it members\/} of the group, though they are technically not
elements of, but rather singleton subsets of, the group considered as a set of points.)
For example, \(\ml{dog}\) is true
of a group just in case each member of the group is a dog.  Not all
two-place predicates are distributive---for example, {\it the boys ate a pizza\/}
does not mean that each boy (separately) ate a pizza.\footnote{See \cite{champollion2020distributivity} for an accessible overview. Also note that a
distributive predicate is one that distributes over {\it plural\/}
individuals.  Distributive predicates are {\it not\/} vacuously true
of the null individual.  In fact,
we assume that every predicate, with the exception of certain
generalized-quantifier predicates, is false if any argument is the
null individual.}

PIP includes a second unselective closure operator, \({\Sigma}x\),
called {\bf summation,\/} which produces a
plurality rather than a truth value.  Like \(\exists\), it binds all
local variables.  The local variables (excluding the
variable of summation $x$, if it is local) are existentially bound, and a 
plurality is formed by taking the union of individuals $x$ that make the
existentially-closed body true.
For example, {\Next[a]} is defined to mean {\Next[b]}, and {\NNext[a]}
can be written out as {\NNext[b]}:

\ex.\a. \({\Sigma}y(\textsc{saw}([x],[y])\wedge\textsc{owns}(z,y))\)\hfill[PIP]
\b. \(\bigcup\{y:\exists x(\textsc{saw}(x,y)\wedge\textsc{owns}(z,y))\}\)\hfill[PC]

\ex.\a. \({\Sigma}z(\textsc{saw}([x],[y])\wedge\textsc{owns}(z,y))\)\hfill[PIP]
\b. \(\bigcup\{z:\exists x\exists y(\textsc{saw}(x,y)\wedge\textsc{owns}(z,y))\}\)\hfill[PC]

\par
Note that \(\exists x\phi\) can be defined in terms of set
abstraction:
\ex. \(\exists x\phi\)\quad is true iff\quad\(\bigcup\{x:\phi\}\neq\varnothing\)

Accordingly, we may take the predicate calculus with set abstraction
and equality to consist of five primitive operations: predicate application, equality, negation,
conjunction, and set abstraction.

Summation allows generalized quantifiers in PIP to act as simple
relations between pluralities, as in {\Next}:
\ex. \a. Every dog barks \\ $\rightsquigarrow\textsc{every}(\Sigma x\, \textsc{dog}([x]), \Sigma x (\textsc{dog}([x]) \wedge \textsc{barks}(x)))$
	\b.\label{most-dogs-bark} Most dogs bark \\ $\rightsquigarrow\textsc{most}(\Sigma x\, \textsc{dog}([x]), \Sigma x (\textsc{dog}([x]) \wedge \textsc{barks}(x)))$

where \(\textsc{every}(x,y)\) iff \((x\subseteq y)\) and
\(\textsc{most}(x,y)\) iff \((|x\cap y|/|x|>\theta)\), with
\(\theta\approx 0.5\).
Notice that the natural language feature of conservativity is
reflected in the repetition of the restriction of the quantifier in
the nuclear scope. This effect is simplified via the use of formula
labels, which we introduce next.

\subsection{Formula labels}\label{sec:formula_labels}

A formula label is a symbol (conventionally, upper-case) that is
defined to be a shorthand for a given formula \citep[cf.\ the \textit{update variables} of][]{keshet2018dynamic}.  We use ``\(\equiv\)''
to define a formula label:
\ex. \(X \equiv \phi\)

The formula in {\Last} defines $X$, but asserts nothing: it is tautologically true.
Once defined, the label can subsequently be used as
a formula, and the effect is exactly the same as inserting the
original formula where the label occurs.  For example, given {\Last}, {\Next[a]} is equivalent to {\Next[b]}.
\ex.\label{X-defas}\a. \(X\vee\neg X\)\hfill[PIP]
\b. \(\phi\vee\neg\phi\)\hfill[PC]

%

Like the other defined constructs of PIP, formula labels are
eliminable, as illustrated in \ref{X-defas}
. In order to expand out
a PIP expression to PC, however, we do require the set of
formula-label definitions, which we write as \(\mathcal A\).  The general
approach is as follows.  Given a PIP expression $\phi$ representing a
discourse, extract the formula label definitions and form the
formula-label assignment function \({\mathcal A}_\phi\), which is
the function such that \({\mathcal A}(X){=}\psi\) just in case 
\(X{\equiv}\psi\)
occurs anywhere in the discourse.\footnote{
The function \({\mathcal A}\) is not well-defined if the discourse
contains \(X{\equiv}\phi\) and \(X{\equiv}\psi\) for some label $X$
and two distinct expressions $\phi$ and $\psi$. If \({\mathcal A}\) is not
well-defined, the discourse is infelicitous:
see Section \ref{sec:felicity} for more discussion.

}
The discourse is then interpreted
(or expanded out into the predicate calculus) with reference to
\({\mathcal A}_\phi\).\footnote{
A consequence of this approach is that
definitions in one sentence may be
used in any other sentence in the same discourse.  That accords with
linguistic facts, when the definition precedes the use, but not when
the definition follows the use.  We return to the issue below, when we
discuss incremental interpretation in Section \ref{sec:felicity}.}

It may be helpful at this point to distinguish between {\it logical equivalence\/}
and {\it annotational equivalence.\/}  We define two formulas to be
annotationally equivalent just in case they have equivalent annotations (that is, the same local
variables, the same formula-label definitions, and---anticipating
Section~\ref{sec:felicity}---the same presuppositions).  If formulas $\phi$ and
$\phi'$ are logically equivalent, they have the same truth value as
each other, in any given context of evaluation.  But to guarantee
that (for example) \(\exists(\phi\wedge X)\) has the same truth value as
\(\exists(\phi'\wedge X)\), logical equivalence of $\phi$
and $\phi'$ is not enough; $\phi$
and $\phi'$ must also be annotationally equivalent.

%
%

%

Generalized quantifiers offer a central example of the use of both
formula labels and summation. A simple example is shown in
\Next:\footnote{A fragment of form ``\(\ldots\wedge X{\equiv}\phi\)'' may
intuitively be read as ``\(\ldots\) where $X$ is defined as $\phi$.''}

\ex.  \a. most $^D$dogs $^B_D$bark.\hfill[LF]
\b.$\textsc{most}(\Sigma x(D \wedge D {\equiv} \textsc{dog}([x])), \Sigma x (B \wedge B {\equiv}(D\wedge \textsc{barks}(x))))$ \hfill[PIP]\label{14b} 

The prefixed superscripts ($D$ and $B$) in \Last[a] mark phrases whose
translations are stored in formula labels. The prefixed $D$ as a
{\it subscript\/} shows that $B$ is subordinate to $D$; this is the
source of the clause ``$D \wedge\ldots$'' in the definition for $B$ in
\Last[b].  Note that {\Last[b]} is logically equivalent to
\ref{most-dogs-bark}, with subordination producing conservativity.

Since formula-label definitions are tautologically true, they can always be
``floated out'' and conjoined at the top level, with no effect on
truth value.  We generally do so for readability, and write for
example {\Next} instead of \ref{14b}:

\ex. $\textsc{most}(\Sigma xD, \Sigma x B)$\\
	\mytab[2ex] $\wedge~ D \equiv \textsc{dog}([x])$
	\mytab $\wedge~ B \equiv(D\wedge \textsc{barks}(x))$

See Section \ref{sec:gq} below for a fuller example of a generalized quantifier in PIP.

Finally, labels also play an important role in the PIP analysis of summation pronouns. For instance, a later pronoun can refer to the dogs that
satisfy $B$ (those that bark).  The next sentence in the discourse might be
{\Next[a]}, which translates as {\Next[b]}:
\ex.\a. they$_{\Sigma d B}$ are loud\hfill[LF]
\b. \(\textsc{loud}({\Sigma}d B)\)\hfill[PIP]

Replacing formula labels with their definitions, {\Last[b]} expands
out to {\Next[a]}, equivalent to the predicate-calculus expression {\Next[b]}:
\ex.\a. \(\textsc{loud}({\Sigma}x
(\textsc{dog}([x])\wedge\textsc{barks}(x))\)\hfill[PIP]
\b. \(\textsc{loud}(\bigcup\{x:\textsc{dog}(x)\wedge\textsc{barks}(x)\})\)\hfill[PC]

The predicate {\sc loud} is distributive, so {\Last} says
that each dog that barks is loud.

\subsection{Worlds}

We assume that the first argument of every predicate is a world, drawn
from a set of worlds ${\mathcal W}\subseteq D$.  We write this distinguished world argument as a subscript on the predicate. (We may sometimes suppress the world
argument, for the sake of simplicity, but it should then be understood.)
An atomic formula \(P_w(x_1,\ldots,x_n)\)
says that, in world $w$, $P$ holds of \(\tp{x_1,\ldots,x_n}\).
We accomplish this technically by means of assumption \ref{2b}, by which the interpretation
function $I$ dispatches to a valuation function $V_w$ indexed by
the world, and then passes the remaining arguments $x_1,\ldots,x_n$ to \(V_w(P)\).

Modals in PIP, analogous to generalized quantifiers, are simply relations between sets of possible worlds. The difference comes in the fact that modals must indicate a \textbf{modal base} \citep{kratzer1981notional}, which determines what flavor of modality is in effect: epistemic, deontic, etc. We make no specific claim about the source of this modal base, but we can represent it as a parameter $\beta$, where $\beta_w$ is a set of worlds accessible from $w$.\footnote{Another option would be to assume an accessibility relation \textsc{r}$_w(u)$ true of any world $u$ accessible from $w$. Then, we could have:
\ex. \a. $\textsc{might}_w^{\textsc{r}}(W_1,W_2)$ \mytab[20ex] $\triangleq~ \textsc{some}(\Sigma u (\textsc{r}_w(u) \wedge u{\in}W_1), W_2)$ 
	\b. $\textsc{must}_w^{\textsc{r}}(W_1,W_2)$ \mytab $\triangleq~ \textsc{every}(\Sigma u (\textsc{r}_w(u) \wedge u{\in}W_1), W_2)$.\z.\z.
} This approach allows us to define existential and universal modals as follows:

\ex.
\a. $\textsc{might}_w^{\beta}(W_1,W_2)$ \mytab[19ex] $\triangleq\, \textsc{some}(\beta_w {\cap} W_1, W_2)$ 
	\b. $\textsc{must}_w^{\beta}(W_1,W_2)$ \mytab $\triangleq\, \textsc{every}(\beta_w {\cap} W_1, W_2)$

The modals take two sets of possible worlds as arguments. The first, usually provided by an \textit{if}-clause, is the modal's quantificational restriction, while the second is its nuclear scope. Notice that while \textsc{some} and \textsc{every} are not dependent on a possible world, $\textsc{might}_w$ and $\textsc{must}_w$ are---hence the subscript $w$.

Since the particular flavor of modality will not be important for the
examples to be discussed, we will generally suppress the $\beta$
parameter, and hence also the $w$ subscript. And in cases without an \textit{if}-clause, we
may suppress the first argument entirely. Essentially, if every world
is accessible and the condition otherwise provided by an \textit{if}-clause is vacuously true, the first
argument to \(\ml{some}\) or \(\ml{every}\) in {\Last} becomes ${\mathcal W}$,
and we accordingly define the one-place versions of the modals as follows:
\(\ml{might}(W)\)
asserts that \(W\) is nonempty and
\(\ml{must}(W)\) asserts that \(W={\mathcal W}\).
For example, we will represent \Next[a] as \Next[b]:
\ex. \a. It might rain.
	\b. $\textsc{might}\,(\Sigma w\, \textsc{rain}_w())$\\
 \mytab[2ex] \(~\leftrightarrow~ \textsc{some}(\Sigma w \top, \Sigma w\, \textsc{rain}_w())\),
 with $\top$ always true\\
 \mytab[2ex] \(~\leftrightarrow~ \Sigma w\, \textsc{rain}_w()\neq\varnothing\).


Given the general presence of a free variable for the world, we should
revisit our earlier assumptions about discourse meaning.
In \ref{discourse_interpretation},%

we defined the meaning of a discourse formed from sentence
meanings \(\phi_1,\ldots,\phi_n\) to be the existential closure of
their conjunction:
\ex. \(\exists(\phi_1\wedge\ldots\wedge\phi_n)\).

However, if all sentence meanings contain $w$ as a free, \textit{external} variable, then
the discourse meaning {\Last} is also an open formula containing free
variable $w$.  
Taking the
presence of world variables into account, then, it is more appropriate
to identify the discourse meaning with the set of worlds that satisfy
the discourse:
\ex.\label{disc_meaning} \({\Sigma}w(\phi_1\wedge\ldots\wedge\phi_n)\).

The discourse is true just in case {\Last} is nonempty.
Notice that {\Last} is simply the proposition (in the set-of-worlds sense) that
satisfies the conjunction of the sentence meanings.
Also, when restricted to worlds consistent with
common-ground knowledge, {\Last} represents the context set produced
by the discourse.

%
%
%

%
%


\subsection{Felicity and presupposition}\label{sec:felicity}

Natural-language sentences not only make
assertions; they also have presuppositions.  To capture presuppositions, PIP
provides expressions of the form \(\phi|\psi\), which
assert $\phi$ and presuppose $\psi$.  Our notation is similar to \posscitet{blamey1986partial} transplication operator\footnote{Except that the transplication operator has the presupposition on the left. See also \cite{beaver2001partial}.}, and may be considered a compact variant of the horizontal-bar notation employed, for instance, by \citet{sauerland2005focus}:
\ex. \(\begin{array}{c}
\phi\\
\hline
\psi
\end{array}\)

%


Unlike Blamey, though, we consider the presuppositions of a formula to be
independent of its truth conditions.
Presuppositions play no role in determining truth: \(\phi|\psi\) is
true iff $\phi$ is true.
Rather, presuppositions determine
whether or not a discourse is {\it felicitous.\/}

There are several ways that a discourse can be infelicitous.  
%
For instance, a discourse is infelicitous if its meaning contains free variables,
making the value indeterminate.  Any free variables in the individual
sentences must be closed by the ``discourse closure''
operator \(\Sigma w\) of \ref{disc_meaning}.  (That is, any free
variables must be local variables).  
%
The discourse is also infelicitous if it contains inconsistent
formula-label definitions or circular formula-label definitions, or if it uses formula labels that are not
defined.  Finally, a discourse is infelicitous if its presuppositions
are not satisfied.  

Let us write \(F\phi\) to mean that expression $\phi$
is felicitous with respect to presuppositions.  (The argument of $F$
may be either a formula or term.)
The final condition on
discourse felicity, then, is that {\Next} must be true:
\ex. \(F\Sigma w(\phi_1\wedge\ldots\wedge\phi_n)\)

\par
We will give a recursive definition of \(F\phi\).  All operators
except the presupposition operator and the primitive PC operators can
be eliminated by expanding them out using their definitions, so we
require recursive clauses only for the presupposition operator and
primitive operators.

Every failure of presuppositional felicity (a false value for an $F$ formula) can be traced to a presupposition violation:
\ex. \(F(\phi|\psi)\) iff \(F\phi \wedge \psi\)

The larger expression is felicitous just in case the body is felicitous and
the presupposition is true.  That is,
if $\psi$ is false, then \(\phi|\psi\) is infelicitous.\footnote{To allow nested presuppositions, this definition could be amended to:
\ex. \(F(\phi|\psi)\) iff \(F\phi \wedge F\psi \wedge \psi\)

}

In a conjunction, we follow \citet{karttunen1974presupposition} in
holding that the first conjunct may satisfy presuppositions of the
second (e.g., \textit{France has a King and the King of France is
  bald}). This is captured in the following definition:\footnote{Note
that this is neither equivalent to strong nor weak Kleene
conjunction. See also \cite{peters1979truth}, who formulates a
three-value truth connective corresponding to Karttunen's theory. Our
resulting system for conjunction is similar in spirit to recent work
by \citet{mandelkern2022witnesses}.} 
\ex. \(F(\phi\wedge\psi)\) iff \(F\phi\wedge(\phi\rightarrow F\psi)\) \label{felix:conj}

For the whole to be felicitous, the first conjunct must be felicitous
outright, but the second conjunct need only be felicitous when the first conjunct is true.

In a set abstraction, the body must be felicitous for every value of
the variable of abstraction:
\ex.\label{F_abstraction} \(F({\bigcup}\{x:\phi\})\) iff \(\forall x\,F\phi\).

%
%
%
And the recursion bottoms out at simple terms, which are always
felicitous:
\ex. \(F\alpha\) is always true, if $\alpha$ is a variable or constant.

For the other primitive predicate-calculus operators, the whole is
felicitous just in case each of the parts is felicitous:
\ex.\a. \(FP(\alpha_1,\ldots,\alpha_n)\) iff \(F\alpha_1\wedge\ldots\wedge F\alpha_n\)
\b. \(F(\alpha=\beta)\) iff \(F\alpha\wedge F\beta\)
\b. \(F\neg\phi\) iff \(F\phi\)

\par
For expressions containing defined operators, felicity is
determined by expanding out the defined operator.  In particular, if
$\phi$ expands to $\phi'$:
\ex. \(F\phi \leftrightarrow F\phi'\).

Doing so for each of the defined operators, and simplifying, one obtains the
following theorems,
which are useful for practical computations.
Writing $x_1,\ldots,x_n$ for the local variables of $\phi$:
\ex.\a. \(F\exists x\phi\) iff \(\forall x F\phi\),
\b. \(F\exists\phi\) iff \(\forall x_1\ldots\forall x_n F\phi\),
\b. \(F\Sigma y\phi\) iff \(\forall y\forall x_1\ldots\forall x_n F\phi\),
\b. \(F\forall y\phi\) iff \(\forall y F\phi\),
\b.\label{F-disjunction} \(F(\phi\vee\psi)\) iff \(F\phi\wedge(\neg\phi\rightarrow F\psi)\),
\b. \(F(\phi\rightarrow\psi)\) iff \(F\phi\wedge(\phi\rightarrow F\psi)\).

\subsection{Applications}\label{sec:applications}
We have now presented the entirety of PIP.
For the remainder of this section, we describe some practical
applications. First, we explore how the felicity conditions just
described may be applied to the problem of sentence-by-sentence
incremental interpretation. Next, we flesh out our discussion of
generalized quantifiers, and show how PIP handles donkey pronouns and
quantificational subordination. We also show how modal subordination
can be treated in parallel to quantificational subordination. %
In these discussions, we mainly concentrate on phenomena covered by
\citet{brasoveanu2007structured,brasoveanu2008donkey,brasoveanu2010decomposing},
since Brasoveanu's system is the closest to ours in terms of
empirical coverage.\footnote{Of course, a good number of other
phenomena have been analyzed via systems extending either
\cite{berg1996some} or \cite{brasoveanu2010decomposing} and making
crucial use of van den Berg's plural information states (sets of
assignments). An anonymous reviewer points out cases like dependent and wide-scope indefinites
\citep{brasoveanu2011indefinites,henderson2014dependent,devries2016independence,kuhn2017dependent},
reciprocals \citep{murray2008reflexivity,dotlavcil2013reciprocals},
and various readings of questions
\citep{dotlacil2020dynamic,roelofsen2023wh}. We leave full
analysis of such phenomena in (systems extending) PIP to future
work. However, we do note here that PIP's formula labels store the
same information as a plural information state, and this at least
points a way to begin capturing these other cases. In particular,
the plural information state corresponding to a label $X$ (whose local variables are $L$) in the context of assignment $g$ is
the set of assignments that verify $X$ and differ from $g$
only with respect to $L$: 
\ex. $\left\{h : g[L]h ~\&~ \svs[h]{X}{=}1\right\}$

}
Finally,
we explain how the two types of terms in PIP (simple terms versus summations) relate to pronouns, with an application to paycheck pronouns. This discussion will lead directly into our analysis of intensional antecedents in Section \ref{sec:analysis}.

\subsubsection{Incremental interpretation}\label{sec:incremental}
One natural question is how one may compute felicity conditions incrementally.
First, we assume that a sentence may only be be felicitously added to
a previously felicitous discourse. This alone already rules out
cross-sentential cataphora for indefinites and formula labels. The
syntactic index on an indefinite is interpreted as a local variable
(a$^d$ dog $\rightsquigarrow$ \(\textsc{dog}([d])\)), while the index
on a pronoun is interpreted as an external variable (it$_d$
$\rightsquigarrow$ \(d\)). Thus, a pronoun {\it it$_d$\/} will be
interpreted as an unbound variable, and hence infelicitous, if the
corresponding indefinite {\it a$^d$ dog\/} occurs in a later sentence. Similarly, if a formula label $X$ is used in a sentence prior to the sentence contributing its definition ``$X{\equiv}\ldots$'', this will also violate the felicity requirements for formula labels.

Supposing then that the subdiscourse consisting of the first $n-1$
sentences is felicitous, what must be true for the
incorporation of the $n$th sentence to yield a felicitous discourse?
Let us write \(\gamma\) for the conjunction of the first $n-1$
sentences, and let us write $\phi$ for the $n$-th sentence.  We
assume:
\ex.\label{hyp} \(F\Sigma w\gamma\),

and we would like to know under what conditions
\(F\Sigma w(\gamma\wedge\phi)\) is true.

For the sake of readability, let us write \(\bm{x} = x_1,\ldots,x_m\)
for the local variables of $\gamma$ (whether or not they appear in $\phi$), and \(\bm{y}\) for the local variables that are unique to $\phi$.
Expanding out \(\Sigma w\) and pushing the $F$ operator inwards, we have:
\ex.\a.\label{32a} \(F\Sigma w\gamma\) iff \(\forall w\bm{x}F\gamma\),
\b. \(F\Sigma w(\gamma\wedge\phi)\) iff \(\forall
w\bm{x}\bm{y}(F\gamma\wedge(\gamma\rightarrow F\phi))\).

Since universal quantification distributes over conjunction and
the variables \(\bm{y}\) do not occur in $\gamma$, the right-hand side
of {\Last[b]} is equivalent to {\Next}:
\ex. \((\forall w\bm{x}F\gamma)\wedge\forall w\bm{x}\bm{y}(\gamma\rightarrow F\phi)\).

But the first conjunct is true by hypothesis \ref{hyp}, given the
equivalence \ref{32a}, so {\Last} is true iff:
\ex. \(\forall w\bm{x}\bm{y}(\gamma\rightarrow F\phi)\).

That is, {\Last} represents the conditions that must hold for the new
sentence to yield a felicitous discourse, given that the discourse up
to now is felicitous.  One can view {\Last} as strict implication for
open formulas: the open formula $\gamma$ entails open formula $F\phi$
just in case the implication holds in every world and for every value
of the free variables.

This result explains, for instance, why the sentence in \Next[b] may not be felicitously added to the (one-sentence) discourse in \Next[a]:
\ex. \a. [Between one and five]$^b$ bunnies live in my yard. 
	\b. \#It$_b$ ate my tulips.

There are some possible worlds in which \Last[a] is true and $b$
denotes a non-singleton plurality.  Since \Last[b] presupposes that
$b$ is is singleton, it is infelicitous in
those worlds.  Hence \Last[a] fails to entail the felicity of \Last[b]
for some values of $w$, making \Last[b] an infelicitous
continuation of the discourse.

\subsubsection{Donkey pronouns and quantificational subordination}\label{sec:gq}
Next let us take a closer look at the PIP representation for
generalized quantifiers by examining a donkey sentence \citep{geach1962reference}, annotated with formula labels:
\ex.\label{donkey} every $^O$[$t^x$ farmer that owns a$^d$ donkey] $^B_O$[$t_x$ beats it$_d$].

The first bracketed subexpression, prefixed with a superscript $O$, is the noun phrase restriction, and
the second, prefixed with a superscript $B$, is the nuclear scope.  We assume \citep[following][]{may1977quantification} 
that syntactic
quantifier-raising has applied, leaving a trace
$t_x$ as the subject of {\it beats.\/}  Nonstandardly, we
assume that a second raising, involving just the determiner, leaves a
trace $t^x$ before {\it farmer;\/} the noun phrase restriction is thus
semantically a sentence, not a predicate.%
\footnote{%
The motivation for quantifier raising is to move the quantifier
(that is, the DP, of type \(\tp{\tp{e,t},t}\) in the Montagovian
analysis) into a position where it has an appropriate argument. Under
our analysis, a determiner is a simple two-place predicate (of type
\(\tp{e,\tp{e,t}}\)), and the $\Sigma$ operator is used to convert
type $t$ expressions to type $e$. We follow \cite{unif} in assuming that 
the DP raises over the nuclear-scope
$\Sigma$ phrase in order to govern the second argument, and the D raises over
the restriction $\Sigma$ phrase in order to govern the first argument. We consider the raising of the D to be a type of head movement, not subject to 
the same island constraints as full DP movement.
\citet{heim1982diss} assumes both DP-- and D-raising, too, for similar reasons.
}
Finally, the prefixed subscript $O$ on
the nuclear scope {\it $t_x$ beats it$_d$\/} indicates that $B$ is
subordinate to $O$, meaning that the restriction
meaning $O$ is incorporated into the nuclear-scope meaning $B$. 
Like the trace, we assume this subscript is the result of the syntactic
movement operation.

The translation of the noun phrase restriction is:
\ex. \({\Sigma}x(O \wedge O\equiv(\textsc{farmer}([x])\wedge\textsc{donkey}([d])\wedge\textsc{owns}(x,d)))\)

Recall that the definition (that is, the \(\equiv\)-expression) is a tautology, so \Last is logically equivalent to \({\Sigma}x O \), which denotes the set of farmers that own a donkey.

The translation of the nuclear scope is:
\ex. \({\Sigma}x(B \wedge B\equiv (O \wedge \textsc{beats}(x,d)))\)

The subordination of $B$ to $O$ is reflected in the inclusion of $O$
in the definition of $B$.  Thus $B$ asserts that $x$ is a farmer and
$d$ is a donkey that $x$ both owns and beats, and {\Last} denotes
the set of farmers that own a donkey and beat it.

Finally, the meaning of the sentence as a whole can be represented as follows, with
the definitions floated out for the sake of readability:
\ex.\label{donkey_meaning} \(\textsc{every}({\Sigma}x O, {\Sigma}x B) \\
	\text{\mytab[3ex]}~~~ \wedge O\equiv (\textsc{farmer}([x])\wedge\textsc{donkey}([d])\wedge\textsc{owns}(x,d))\\
	\text{\mytab}~~~ \wedge B\equiv (O \wedge \textsc{beats}(x,d))
\)

The predicate \textsc{every} takes two plural individuals and
asserts that the first (the set of farmers that own a donkey)
is a subset of the second (the set of farmers that own a donkey and
beat it). 

A key point to observe is the licensing of the pronoun {\it it$_d$,\/}
which translates as the unbracketed variable $d$.  An unbracketed
variable $d$ is infelicitous unless there is a bracketed occurrence
$[d]$ within the scope of the same closure operator. (An external
variable other than $w$ is infelicitous, and an unbracketed variable
is external unless it co-occurs with a bracketed instance of the same
variable.) In the clause \(O\wedge\textsc{beats}(x,d)\), the $d$
provided by {\it it$_d$\/} is rendered felicitous by the bracketed variable
\([d]\) occurring within $O$, as becomes evident if one
expands out $O$ using its definition. (The $x$ provided by the trace
of {\it farmer\/} is similarly licensed by an occurrence of \([x]\)
within $O$.)

%
%

Just as the restriction of a quantifier is incorporated into its nuclear scope, the nuclear scope of a previous quantifier may be incorporated into the restriction of a later quantifier, a phenomenon known as quantificational subordination. For instance, our donkey discourse may continue as \Next[a], which translates as \Next[b], with the label definition floated out again for readability:
\ex.\label{qsubord_null} \a. most $\tp{}\!_B\,$ $^T_B$[$t_x$ treat it$_d$ well otherwise] \hfill [LF]
	\b. $\textsc{most}(\Sigma x B, \Sigma xT)$\hfill [PIP]\\$\text{\mytab[1ex]}~~~\wedge T\equiv(B \wedge \textsc{treat-well}(x,d))$ 

Notice that the (implicit) restriction of \Last is provided by the nuclear scope
label $B$ of \ref{donkey}, to which the nuclear
scope label $T$ is then subordinated, just as the
nuclear scope in \ref{donkey} was subordinated to its restriction.  This yields the correct
interpretation: most of the donkeys that are owned and beaten are
donkeys that are owned, beaten, and otherwise treated well.

Note the difference between the licensing of the pronoun
{\it it$_d$\/} in this case, versus in the case of simple
cross-sentential anaphora \ref{xsent}.  The bracketed occurrence
\([d]\) provided by {\it a dog\/} is a local variable in the first
sentence of \ref{xsent}, but in \ref{donkey_meaning}, the \([d]\)
provided by {\it a donkey\/} is embedded under a closure operator.
Thus ``it$_d$ brayed'' would {\it not\/} be felicitous as a
continuation after \ref{donkey_meaning}; that is empirically
correct. The subordination that occurs in \ref{qsubord_null} provides
access to an antecedent \([d]\) that would not otherwise be
accessible.

\subsubsection{Modal subordination}
Modal subordination is treated precisely parallel to quantificational subordination. Since the restriction of a modal is usually provided by an \textit{if}-clause, let us first consider a conditional donkey sentence (\textit{If Sal has a pet, it must be a donkey}). We assume that modals scope over their prejacent:
\ex. \a. If $^P$[Sal has a$^x$ pet], must $^D_P$[it$_x$ be a donkey]\hfill [LF]
	\b. $\textsc{must}_w(\Sigma wP, \Sigma w D)$\hfill [PIP]\\
		$\text{\mytab}~~\wedge P\equiv \textsc{pet-of}_w([x],\textsc{Sal})$\\
		$\text{\mytab}~~\wedge D\equiv (P \wedge \textsc{donkey}_w(x))$

Just as with individual quantifiers, the nuclear scope of a modal is
subordinate to its restriction (\textit{if}-clause). This is indicated
by the subscript prefix $P$ in \Last[a], and the ``$P \wedge\ldots$''
in the definition for $D$ in \Last[b]. This subordination gives
the pronoun \textit{it} in the nuclear scope access to
an antecedent indefinite in the restriction.

Next, here is an example of modal subordination \citep{roberts1987diss}:
\bigskip
\ex. \a. might $\tp{}$ \mytab[12ex]$^W$[a$^x$ wolf enter]\hfill \textit{A wolf might enter} \label{wolf}
	\b.  would $\tp{}\!_W\,$ \mytab$^E_W$[it$_x$ eat Tasty Tim] \hfill \textit{It would eat Tasty Tim} \label{tim}
	
\ex. \a.$\textsc{might}_w(\Sigma w W)$ \mytab[22ex] $\wedge ~W\equiv (\textsc{wolf}_w([x]) \wedge \textsc{enters}_w(x))$
	\b. $\textsc{must}_w(\Sigma w W, \Sigma w E)$ \mytab $\wedge ~E \equiv(W \wedge \textsc{eats}_w(x,\textsc{Tim}))$
	
The first sentence, in \ref{wolf}, has a null restriction, which we
take to be a tautology and suppress in \Last[a] for readability. The
nuclear scope of \ref{wolf} is stored in the label $W$. The second
modal, in \ref{tim}, is subordinate to the first. Its restriction is
also null, but this null element is anaphoric to the nuclear scope
of \ref{wolf}, as indicated by the index $W$. (Note that this anaphoric
null anaphor is identical to the one in the
quantificational-subordination example \ref{qsubord_null}.)
The nuclear scope of \ref{tim} is subordinate to its restriction, as
usual, and hence the nuclear-scope meaning $E$ includes $W$,
giving the pronoun in the nuclear scope access to the
antecedent that occurs in the preceding sentence.


\subsubsection{Simple, summation, and paycheck pronouns}
PIP has two kinds of terms: simple terms, like single lowercase variables, and summation terms. These correspond to the two types of terms in our definition of predicate calculus: simple terms and set abstractions. (Recall that this definition is not more complex than the usual one, since quantification is defined in terms of set abstraction.) Summation terms are an essential component of the PIP analysis for generalized quantifiers, allowing the quantifiers themselves to denote simple, two-place predicates over individuals. 

With two kinds of terms, it stands to reason that either one could be the meaning of a pronoun, and this is exactly what we find. PIP represents many pronouns with simple terms, but some, such as summation pronouns, denote summation terms. These two representations make predictions, beyond the obvious differences in meaning. In particular, summation pronouns are exhaustive in a way that simple pronouns are not. Consider the following examples:
\ex. \label{somegirls}\begin{samepage}\a.\label{somegirlsa} Some$^g$ girls were having lunch in the cafeteria. \b.\label{somegirlsb} They$_g$ waved to some other girls having lunch there, too.\z.
\end{samepage}

\ex. \label{mostgirls} \a. Most$^g$ girls $^L$[$t_g$ were having lunch in the cafeteria]. \b. \#They$_{\Sigma gL}$ waved to some other girls having lunch there, too.

As in typical dynamic systems, PIP translates indefinites as simple
variables, such as the variable $g$ in \ref{somegirls}. Such
terms may serve as the denotation for a later pronoun, as shown. The
variables introduced by generalized quantifiers are bound inside the
quantification, though, and cannot serve as the denotation for a later
pronoun. Instead, any reference to the individuals quantified over,
such as the girls eating lunch in \ref{mostgirls}, is necessarily via
summation terms, as shown: \textit{they} denotes the sum of girls
satisfying the nuclear scope of the quantifier \textit{most}.

This distinction makes a notable predication about the exhaustivity of
the pronoun with respect to its intuitive ``antecedent description.'' In the examples
{\LLast} and {\Last}, what we mean by the intuitive antecedent description is the set
of girls having lunch. That is the set that arises as the denotation
of the nuclear scope under the traditional analysis of
\ref{somegirlsa} as a generalized quantifier. The variable $g$ in \ref{somegirlsb}
may denote any plurality of girls having lunch, which may be
the entire set of girls having lunch, but need not be. In particular, it is fine to mention
\textit{other girls} in \ref{somegirlsb}, indicating that \textit{they}
does not necessarily denote the complete set of girls having
lunch.%
\footnote{%
Providing a rigorous semantic account of {\it other\/} would take us
too far afield, but, whatever the account,
it must have the consequence that the denotation of
{\it some other girls\/} is disjoint from the denotation of {\it they,\/}
and that infelicity results if {\it they\/} exhausts the set of
relevant girls, forcing the denotation of {\it some other girls\/} to
be the empty set.
} 
In our analysis, this fact obtains because there is no
generalized quantification involved. Rather, the first sentence
asserts explicitly that $g$ is a plurality of girls having lunch, an assertion
that does not entail exhaustiveness.

Summation pronouns, on the other hand, must be
exhaustive relative to their antecedent description. Since
\textit{they} in \ref{mostgirls} is a summation
pronoun, it denotes the complete set of girls having lunch in the
cafeteria. This explains why it is odd to mention other girls having
lunch there in \Last[b]. In our analysis, this is because {\Last[a]}
{\it does\/} involve a generalized quantifier, and the summation
pronoun does refer back to the nuclear-scope meaning, denoting the
(complete) set of elements that satisfy the nuclear scope.

Note that the exhaustiveness holds for pronouns that refer back to antecedents within intensional contexts, as
well. Consider:

\ex.\a. There must be some animals in the shed.
\b. \#They are bothering the other animals in the shed.

We analyze {\it they\/} in {\Last[b]} as a summation pronoun, which must refer
exhaustively to the animals in the shed, leaving nothing for ``the
other animals'' to refer to.

A similar set of examples based on one due to
\cite{evans1980pronouns} is given below:

\ex. \a. A few senators that I know like Kennedy. \b.But they don't trust the other senators that I know who like Kennedy.

\ex. \a. Few senators that I know like Kennedy. \b. \#But they don't trust the other senators that I know who like Kennedy.

We are not alone, either, in observing two different kinds of pronouns
along these lines. This assumption is very common across the entire
E-type literature. One particularly similar analysis, for instance, is
due to \citet{rooy2001exhaustivity}, who proposes two types of
pronouns, quite similar to our simple and summation types.
He posits a class of \textit{descriptive pronouns}, which denote ``the exhaustive set of individuals denoted by the description recovered from the clause in which [an] antecedent occurs.'' These contrast with the class of \textit{referential pronouns}, which are not exhaustive in the same way.\footnote{These pronouns, van Rooij suggests, are made exhaustive via a connection to the speaker's intended referents.}

Summation pronouns are also key to the PIP analysis of paycheck pronouns \citep[following][]{keshet2018dynamic}:
\ex. \a. Almost every$^x$ girl brought in the $^P$[$t^p$ paper she$_x$ wrote].
	\b. Few$^x$ of them forgot it$_{\Sigma p P}$ at home.

The definite description in \Last[a] stores its NP predicate in the label $P$, giving the pronoun in \Last[b] the value \Next:
\ex. \a. $\Sigma p P~~$ where $P\equiv (\textsc{paper}([p]) \wedge \textsc{wrote}(x,p))$\hfill [PIP]
	\b. $\bigcup \{p : \textsc{paper}(p) \wedge \textsc{wrote}(x,p)\}$\hfill [PC]
	
Notice that $x$ is an external variable in the value for $P$ and
therefore $x$ remains free in the expression $\Sigma p P$. This allows
$x$ to be bound by another, higher operator, in this case the
generalized quantifier \textit{few}. Thus, the \textit{it} in
\LLast[b] can refer to papers not mentioned before: those written by
the few forgetful students. 
In general terms, the antecedent description contains a
free variable that is bound by
one operator where the original description occurs, and by a different operator 
where the paycheck pronoun occurs. That is the defining feature of a paycheck
pronoun. It%
 will also figure prominently in
our analysis of intensional antecedents to anaphora. 

Finally, the same features that allow for paycheck pronouns are key to the PIP analysis of strong donkey pronouns. The translations given so far yield only weak donkey pronouns \citep[cf.][]{schubert1989generically}: each farmer need only beat one of their donkeys to satisfy \ref{donkey_meaning}, for instance. \cite[Section 2.7]{unif} details how to capture the strong version. 
The antecedent indefinite (e.g.\ \textit{a donkey}) is treated as a generalized quantifier, rather than a simple indefinite, thus generating a formula label storing ``$\textsc{donkey}([d]) \land \textsc{owns}(x,d)$''. This formula may serve as the antecedent to a summation pronoun in the nuclear scope, where $x$ is bound locally (like a paycheck pronoun), yielding all the donkeys that $x$ owns:
\ex. \begin{samepage}\(\textsc{every}(\Sigma xF, \Sigma xB)\)
	\a.[] \(\land~F \equiv (\textsc{farmer}([x]) \land \textsc{some}(\Sigma dD, \Sigma dO))\)
	\b.[] \strut\hskip 6ex\(\land~~ D \equiv \textsc{donkey}([d])\)
	\b.[] \strut\hskip 6ex\(\land~~ O \equiv (D \land \textsc{owns}(x,d))\)
	\b.[] \(\land~B \equiv(F \land \textsc{beats}(x,\Sigma dO))\)
	\z.
	\end{samepage}

\Last is true only when each farmer beats all their donkeys.
Last, \citet[Section 3.6]{keshet2018dynamic} shows how a quite similar system can generate the same sorts of mixed weak and strong readings as \cite{brasoveanu2008donkey}, who also encapsulates the weak/strong difference in two different versions of indefinites.

\section{Analysis} \label{sec:analysis}
The crux of our analysis for Stone's intensional anaphora cases is that they involve a paycheck pronoun whose free variable is the world variable itself. Ignoring pronoun presuppositions for now, we get the formula \NNext[a] for the discourse in \Next, which is logically equivalent to \NNext[b] once formula labels are eliminated:
\ex. might $^X$[there be a$^x$ winner]. she$_{\Sigma xX}$ is a woman.\hfill [LF]

\ex. 
\a. $\Sigma w\left(\begin{array}{l}
	\textsc{might}_w\!\left(\big.\Sigma w (X\wedge X \equiv \textsc{winner}_w([x]))\right)\\ 
	~~\wedge ~\textsc{woman}_w(\Sigma x X)
	\end{array}\right)$\hfill [PIP]
\b. $\Sigma {\color{red}w} \left(\big.\textsc{might}_{\color{red}w}(\Sigma {\color{Cyan}\bm w}\, \textsc{winner}_{\color{Cyan}\bm w}([x])) \wedge \textsc{woman}_{\color{red}w}(\Sigma x\, \textsc{winner}_{\color{red}w}([x]))\right)$

Notice that when $X$ is first used, inside the prejacent to \textsc{might}, the world variable $w$ is bound by the local summation $\Sigma w$ inside the scope of \textsc{might}. (This is reflected by the shared blue color in \Last[b].) This, then, is a \textit{de dicto} world, one occurring inside the intensional context created by \textit{might}. However, $w$ occurs as an external variable within the formula $X$, so when $X$ appears for a second time, as part of the summation pronoun \textit{she}, it is not bound by the summation $\Sigma x$. Instead, it is bound by the $\Sigma w$ atop the entire discourse, as reflected by the shared red color in \Last[b]. Thus, $\Sigma w X$ is the set of (de dicto) worlds $w$ where there is a winner in $w$, but $\Sigma x X$ is the set of winners in the (de re) ``real'' world at the discourse level. 

Notice that world variables are never
local
in our system: there is never any expression involving `$[w]$'. Thus, the world associated with a predicate must come from the
context%

of evaluation, whether it is the discourse world, or a world shifted by some intensional operator. Either way, this explains why the referent(s) of a summation pronoun will never come from multiple worlds: the predicate subscripts $w$ inside the summation pronoun must all share a single value across the summation, one single world.

The other elements of the analysis are pronoun presuppositions and
felicity conditions. We assume that all pronouns presuppose that their
denotations are non-empty, with singular pronouns further presupposing
that their denotations are singular. This is sketched in
\Next:\footnote{In keeping with our assumption that the first argument
of every predicate is a world, we should technically write
\(\textsc{single}_w(\tau)\). However, $w$ has no effect on the value, so we suppress it.}
\ex. \textit{she}$_\tau$ $~\rightsquigarrow~ \tau | \textsc{single}(\tau)$ \hfill where $\textsc{single}(\tau) \leftrightarrow |\tau| {=} 1$

(The metavariable $\tau$ is intended to range over either type of
term, simple variables or summations, indiscriminately.) Therefore, the felicity conditions for a sentence with a singular pronoun will always involve the uniqueness of the pronoun's denotation.

Finally, let us recall, from Section \ref{sec:incremental} above, the presuppositional felicity conditions for incremental addition of a new sentence $\phi$ to an existing felicitous discourse $\gamma$ (where $\bm{x}$ represents the combined local variables of $\gamma$ and $\phi$):
\ex. \(\forall w\bm{x}(\gamma\rightarrow F\phi)\) \label{incfel}

For a sentence $\phi$ with a singular pronoun denoting $\tau$ and no other presuppositions, this reduces to:
\ex. \(\forall w\bm{x}(\gamma\rightarrow \textsc{single}(\tau))\)

In other words, in order for $\phi$ to be felicitous after $\gamma$, $\gamma$ must imply that $\tau$ is singular in all worlds and under all valuations of local variables. This condition will fail for $\tau$ whose values are empty in certain worlds consistent with $\gamma$ (i.e., certain worlds in the context set).

\subsection{Possible burgers}
Let us begin our analysis with the following little discourse:
\ex.\a. Andrea is eating a cheeseburger.
\b. It is large.

{\Last[a]} translates as:

\ex. $\textsc{burger}_{{w}}([b]) \wedge \textsc{single}(b)\wedge \textsc{eats}_{{w}}(\textsc{andrea},b)$\label{eats}

The world variable \({w}\) is an external variable (that is, an
unbracketed free variable), but in the larger context of the discourse
it is bound by the implicit ``\(\Sigma w\)'' at the
beginning of the discourse.  

{\it It is large\/} translates as \Next, with presuppositional felicity conditions \NNext:

\ex. \(\textsc{large}_{{w}}(b|\textsc{single}(b))\)\label{large}

\ex. $F(\textsc{large}_{{w}}(b|\textsc{single}(b)))$, which simplifies to $\textsc{single}(b)$

Given \ref{incfel}, \ref{large} will only be felicitous after \ref{eats} in case the following holds:
\ex. $\forall w\forall b\left(
\begin{array}{l}
	(\textsc{burger}_{{w}}(b) \wedge \textsc{single}(b)\wedge \textsc{eats}_{{w}}(\textsc{andrea},b))\\ 
	~~~\rightarrow \textsc{single}(b)
\end{array}\right)$

Since this formula is clearly true, and there are no unbound variables or formula labels, the sentence is felicitous in this context. In particular, the assertion ``$\textsc{single}(b)$'' in \ref{eats} satisfies the identical presupposition in \ref{large}.

Now let us consider the same discourse, but with the first sentence
embedded under {\it might.\/}

\ex.\label{might}\a. Andrea might be eating a cheeseburger.\label{might-a}
\b. \#It is large.\label{might-b}

{\Last[a]} translates as follows:

\ex. \a. might $^E$[Andrea be eating a$^b$ cheeseburger]\hfill [LF]
	\b. $\textsc{might}_w(\Sigma wE)$\hfill [PIP]\\
	\mytab[2ex]$~~~ \wedge~ E\equiv (\textsc{burger}_{{w}}([b]) \wedge \textsc{single}(b)\wedge \textsc{eats}_{{w}}(\textsc{andrea},b))$

Expanding out {\sc might} and $E$, the latter
is equivalent to:

\ex. \(\exists w (\textsc{burger}_{{w}}([b]) \wedge \textsc{single}(b)\wedge \textsc{eats}_{{w}}(\textsc{andrea},b))\)

Since the variable $b$ is local to the summation $\Sigma w E$
in {\LLast[b]}, it cannot be used felicitously as the value for a pronoun at the top, discourse level. Therefore, only summation pronouns may target this variable, as shown in the following:
\ex. \a. It$_{\Sigma bE}$ is large\hfill [LF]
	\b. $\textsc{large}_w(\Sigma b E \,|\, \textsc{single}(\Sigma b E))$\hfill [PIP]

Finally, the felicity conditions of the new discourse are as follows (where $E$ is as defined above):
\ex. $\forall w(\textsc{might}_w(\Sigma w E) \rightarrow \textsc{single}(\Sigma b E))$

Now, the set of possible worlds includes those where, say, Andrea is
fasting. In those worlds $\Sigma b E$ is empty, since there are no
cheeseburgers Andrea is eating, and therefore $\textsc{single}(\Sigma
b E)$ will be false. But among worlds in which Andrea is fasting,
there are some in which we do not {\it know\/} that she is fasting,
and in which it is thus true (under the epistemic sense of {\it might\/})
that Andrea \textit{might} be eating a
cheeseburger. Therefore \Last comes out as false, due to the
worlds in which Andrea might be---but isn't---eating a cheeseburger.

Similar reasoning holds for \textit{they are large}, which presupposes a cardinality of burgers of greater than one, in place of the singleton cardinality presupposition of \textit{it}. In either case, the pronoun presupposes that its value is non-empty, leading to presupposition failure in worlds that yield an empty value. (Let us call this the \textbf{existential presupposition} of pronouns.) For completeness, the summation interpretation of \textit{they are large} is given in \Next:

\ex. $\textsc{large}_w(\Sigma b E \,|\, \textsc{plural}(\Sigma b E))$\hfill where $\textsc{plural}(\tau) \,\leftrightarrow\, |\tau|>1$


Now, in certain cases of presupposition failure, a
process of accommodation \citep{lewis1979scorekeeping} can
implicitly alter a context set incompatible with a presupposition into
a minimally different, compatible context set. For instance,
one might implicitly assert the content of the presupposition of {\Last}, 
causing assignments where Andrea is eating the wrong number of burgers to
be removed from the context set.  If this accommodation went through, then, the
discourse in \ref{might} would be felicitous, contrary to fact. 

As many researchers have pointed out, though, the existential presupposition of pronouns is a strong one, not subject to accommodation. For instance, \citet{tonhauser2013projective} conclude that a pronoun imposes what they term a ``strong contextual felicity constraint'' that the pronoun has a referent. This means, essentially, that there is no way around this type of presupposition---it must be supported by the context in order for its containing expression to be felicitous.

\subsection{Possible candidates}
Next, let us consider the case where the individuals referred to under a modal actually exist, even though the predicate used to refer to them may be empty in the actual world. Recall that this case was problematic for the Stone/Brasoveanu-style presupposition simply requiring that a pronoun's referent exist in the world of evaluation.
\ex. \a. There may already be a winner in the mayoral race. 
	\b. \#She is a woman, you know.\label{she-woman}
	
By way of reminder, the scenario involved is as follows: a known set of (real-life) candidates have run for mayor, with no write-in candidates allowed. Late on election night, the speaker has not heard any news, but figures the tabulation might be over and a winner declared. Even so, and even if the speaker only considers one candidate as a possible winner, \Last[b] is odd. In other words, \Last[b] cannot mean \textit{the likely/possible/potential winner is a woman}.

This result follows from the full PIP analysis of \Last in \Next, with felicity conditions in \NNext (we omit the singular assertion of \textit{a winner} for readability):

\ex. $\textsc{might}_w(X \wedge X\equiv\textsc{winner}_w([x])) \wedge \textsc{woman}_w(\Sigma x X | \textsc{single}(\Sigma x X))$

\ex. $\forall w\left(\big.\textsc{might}_w(\Sigma w\,\textsc{winner}_w([x])) \rightarrow \textsc{single}(\Sigma x\,\textsc{winner}_w([x]))\right)$

Once again, there will be worlds with no winner actually declared yet, but where (for all we know) there \textit{might} be a winner declared. In these worlds, $\Sigma x\,\textsc{winner}_w([x])$ will be empty, not singular, rendering \Last false. It (correctly) makes no difference under this analysis whether the potential winners exist in the real world or not.

\subsection{Possible animals}
Let us turn next to universal modals. 
%
%
%
We assume, as standard, that epistemic \textit{might} and \textit{must} are realistic; that is, they include their own world of evaluation in their sets of accessible worlds. With this in mind, consider the \textit{must}-sentence discourse in \Next, with its translation in \NNext:
\ex. \a. There must be some sort of animal in the shed.\label{shed}
	\b. It's making quite a racket!\label{racket}

\ex. $\textsc{must}_w(\Sigma w X) \land \textsc{loud}_w(\Sigma x X \,|\, \textsc{single}(\Sigma x X))$\\$\text{\mytab[2ex]}~~\land X\equiv (\textsc{animal}_{w}([x]) \land \textsc{single}(x) \land \textsc{in-shed}_{w}(x))$
	
The felicity conditions for adding \ref{racket} after \ref{shed} are as follows:
\ex. $\forall w\left(\textsc{must}_w(\Sigma w X) \rightarrow \textsc{single}(\Sigma x X)\right)$


Since \ref{shed} asserts that all accessible worlds are ones with an
animal in the shed---which, by a standard scalar implicature, is
unique---and since the world of evaluation is one of the accessible
worlds, it follows that there is a single animal in the shed in the world of
evaluation after the assertion of \ref{shed}. In any world $w$ where
$\textsc{must}_w(\Sigma w X)$ is true, $\Sigma x X$ will be the singleton animal
in the shed in $w$. This satisfies the felicity conditions \Last.
%

There are two points of note concerning the Stone/Hardt/Brasoveanu
analysis of this case. First, as noted above, for any given world of
evaluation $w$, not every possible animal in a world accessible from
$w$ will exist in $w$. In other words, of all the possible animals
that might be in the shed, only a subset will exist in the actual
world. Therefore, the presupposition that these individuals must exist
in $w$ is not met, and the pronoun is predicted, incorrectly, to be
infelicitous. Second, even if this presupposition were met, the
pronoun's value would be plural in Brasoveanu's system: it would
represent all possible animals in the shed, summing across all worlds. There would be no way to
get from this full set of animals to just the single animal in the shed in
the real world. 

\subsection{Possible witches and bathrooms}
Finally, we present two further scenarios from the previous literature on intensional anaphora, which receive straightforward analyses using PIP:
\ex. Hob thinks a witch has blighted Bob's mare, and Nob wonders whether
she (the same witch) killed Cob's sow. \citep{geach1967intentional}\label{hobnob}

\ex. \a. Either John doesn't own a donkey, or he keeps it very quiet. \citep{evans1977pronouns1}
	\b. Either there's no bathroom in this house or it's in a funny	place. \citep[attributed to Barbara Partee in][]{roberts1987diss} \label{bathroom}

Much has been written about Geach's Hob/Nob sentence in \ref{hobnob}. (See \citealt{ninan2022hobnob} for a recent overview.) Unsurprisingly, PIP predicts an analysis for \ref{hobnob} dubbed ``descriptivist'' by \citet{ninan2022hobnob}.
The PIP translation of the first clause (Hob's thought) will contain a clause like the following:
\ex. $X\equiv(\textsc{witch}_{ w}([x]) \land \textsc{blighted-mare}_{ w}(x))$

Now, if the second clause (Nob's wondering) treats \textit{she} as a
summation pronoun whose antecedent description is the clause labeled $X$,
\textit{she} will refer to the witch who blighted Bob's mare {\it in Nob's
belief worlds.\/}
Given the presupposition that the pronoun's referent exists and is
unique, the pronoun will only be felicitous in a context where every
accessible world has one such a witch. This implicates that Nob's
belief worlds are the same as Hob's at least with respect to the
existence of the witch, a state of affairs that is consistent with
the medieval context of the sentences.\footnote{It is reasonable to
ask why an analogous implicature does not rescue \ref{might-b}.  In
the case of \ref{might-b}, the necessary implicature would
take {\it might\/} to mean {\it must,\/} in violation of the usual scalar
implicature by which {\it might\/} implicates that {\it must\/} is false.}
(Geach describes the relevant scenario as a town with an ``outbreak of witch mania.'')
If so, there is also one relevant witch in each of Nob's belief
worlds, providing a proper antecedent for {\it she.\/}

Now, certain authors (including Geach) have proposed that \ref{hobnob}
may be true even if Nob is unaware of Hob and Bob---say if both Hob
and Nob read the same (untrue) rumors in the newspaper about a
particular witch running amok in their town \citep[see][Example
  2]{edelberg1986new}. To the extent that this is true, we propose in
that case treating the indefinite \textit{a witch} as a so-called
wide-scope indefinite
\citep{fodor1982indefinites,kratzer1998pseudoscope}. For instance, the
word ``certain'' sounds fine with the indefinite in this case, and the
particular value for the indefinite can be added, as follows:
\ex. Hob thinks a (certain) witch has blighted Bob's mare---namely, the witch he read about in the newspaper.

Furthermore, if the newspaper article mentions multiple witches
ruining the town, \ref{hobnob} is no longer acceptable: it sounds odd
for Nob to be wondering about the same witch, if he has no knowledge
of Hob or Bob. To sketch the outline of an analysis of wide-scope
indefinites, we may take the description
\(\textsc{witch}_w([x])\) to be moved out of the scope of {\it believes\/}
and into the scope of a contextually-provided intensional operator
(such as ``newspapers claim'') with a body labeled $Y$---that is, \((Y\equiv\ldots\textsc{witch}_w([x])\ldots)\).
What remains inside the scope
of Hob's beliefs is \(Y\wedge\textsc{blighted-mare}_w(x)\). Then $Y$
provides an antecedent description suitable
for a later summation pronoun {\it she\(_{\Sigma xY}\)\/} in Nob's
attitude ascription.

Next, let us turn next to the translation of \ref{bathroom}:
\ex. $(\lnot\exists X \vee \textsc{funny-place}(\Sigma b X\,|\,\textsc{single}(\Sigma b X)))$\\$\text{\mytab[2ex]}~~\wedge X\equiv(\textsc{bathroom}([b]) \land \textsc{here}(b))$

As described in Section \ref{sec:felicity}, the felicity conditions for disjunction are as follows repeated from \ref{F-disjunction}:
\ex. \(F(\phi\vee\psi)\) iff \(F\phi\wedge(\neg\phi\rightarrow F\psi)\)

Since $\lnot \exists X$ contains no presuppositions, this translates to \Next for \LLast:
\ex. $\exists b\, \textsc{bathroom}(b) \rightarrow \textsc{single}(\bigcup\{ b : \textsc{bathroom}(b)\})$

This condition goes through, as predicted, if
the use of the singular {\it no bathroom,\/} as
opposed to {\it no bathrooms,\/} creates a scalar implicature that there is at most one bathroom.\footnote{It would also go through in cases with multiple bathrooms, if \textit{it} acts like a strong donkey pronoun. See \cite{unif} for a compatible treatment of strong donkey anaphora.}
%

\section{Previous analyses} \label{sec:prev}
In this section, we discuss previous literature which addresses intensional anaphora. First, we detail the IP-CDRT system due to \cite{brasoveanu2010decomposing}, as a representative value-based approach. Next, we examine some more recent, hybrid approaches more similar to the one presented here.

\subsection{Stone, Stone \& Hardt, and Brasoveanu}\label{intro prev}
Although they differ in many respects, the systems presented in 
\cite{stone1999reference,stone1999dynamic} and \cite{brasoveanu2010decomposing} 
share some crucial aspects in common. For the sake of concreteness, we will 
sketch the account that \cite{brasoveanu2010decomposing} gives within the
IP-CDRT system, %
and note when we do not believe the criticism applies equally to \cite{stone1999reference} and
\cite{stone1999dynamic}.  


The states in IP-CDRT, following \cite{berg1996some}, 
are sets of assignments, often represented as tables in which columns
correspond to variables and rows correspond to assignments.  
States keep track of what is
known so far in the discourse; an assignment \(g\) in the current
state encapsulates one way of assigning values to variables that
satisfies the discourse so far. One specific column in each IP-CDRT table, which we will label 
$w^*$, contains the same value in every row, and represents the top, discourse-level world of evaluation for that table.
We will follow the convention of using uppercase 
variables $G$, $H$, and $K$ for tables, and lowercase $g$, $h$, $k$ for single 
assignments.

Lexical predicates in Brasoveanu's system  are interpreted 
distributively 
relative to a table; and distributive application is represented by using curly braces around 
the arguments, instead of the usual parentheses. The world of evaluation is 
indicated as a subscript. That is, relative to a model $\mathcal M$:
\ex. $\textsc{eat}_{w}\{x,y\}$ \hfill  ``he$_x$ is eating$_w$ it$_y$''
	\\ $\rightsquigarrow \lambda G\,.\,\forall g{\in}G  \left(\textsc{eat}_{g(w)}(g(x),g(y))\right)$
	\\ = $\lambda G\,.\,\forall g{\in}G\left(\tp{g(x),g(y)} \in {\mathcal M}(\textsc{eat}, g(w))\right)$

%
The formula \(\textsc{eat}_w\{x,y\}\) is true, relative to a table like 
\Next, provided that each row contains an $x$ that ate $y$ in world $w$ 
according to $\mathcal M$:
\ex. \begin{tabular}{c|c|c}
$w$ & $x$ & $y$\\\hline
World 1 & Person 1 & Burger 1\\
World 1 & Person 2 & Burger 2\\
World 2 & Person 1 & Burger 2\\
$\vdots$ & $\vdots$ & $\vdots$\\
\end{tabular}\label{eat table}

\par
Modals and other intensional operators introduce new world-denoting 
variables, against which embedded predicates are evaluated. A heavily
simplified version of IP-CDRT modals is sketched in {\Next}, using
notation defined in {\NNext}:
\ex.\a. $\textbf{might}_{w_0, a_0}^{w_1} \phi$\hfill (epistemic)\\
$\rightsquigarrow\lambda G\,.\,\textbf{epis}_{w_0}(a_0)(G) \land 
\textbf{max}^{w_1}(\phi)(G) \land G(a_0) {\cap} G(w_1) {\neq}\varnothing$
	\b. $\textbf{must}_{w_0, a_0}^{w_1} \phi$\hfill (epistemic)\\
	$\rightsquigarrow\lambda G\,.\,\textbf{epis}_{w_0}(a_0)(G) \land 
	\textbf{max}^{w_1}(\phi)(G) \land G(a_0) {\subseteq} G(w_1)$

\ex. Definitions (in simplified form):\begin{samepage}
	\a. $G(x)$ \mytab[10ex] $\triangleq \{g(x) : g{\in}G\}$
	\b. $\textbf{epis}_w(x)$ \mytab $=\lambda
        G\,.\,G(x)=\{w':\textsc{epistemic}_{g(w)}(w')\}$
	\b. $\textbf{max}^x \phi$ \mytab $=\lambda G\,.\,\sv{\phi}(G)
	\land \lnot\exists H \left(\big.\sv{\phi}(H) \land G(x){\subsetneq}H(x)\right)$
	\z.\end{samepage}

According to the definitions above, a modal compares two columns: $a_0$ is the set of worlds accessible (epistemically, say) from the evaluation world $w_0$, and $w_1$ is the maximal set of worlds satisfying a given formula $\phi$. The column values in $G$ for these variables, namely $G(a_0)$ and $G(w_0)$ must stand in the correct relation for the modal in question: subset for \textbf{must} and overlapping for \textbf{might}.

For example, the %
table in \Next[b] is one of several which make the formula in \Next[a] is true: the $a_0$ column holds the worlds accessible from World 1, the $x$ column holds burgers which exist in the corresponding world in the $w$ column. The $w_1$ column is a partial copy of the $a_0$ column: the $\star$ is a dummy value, allowing us to represent accessible worlds in the $a_0$ column where there is no (salient) burger, and hence no value in the $w_1$ or $x$ columns.%
(The clause $\qv{[x]}$ is part
of the contribution of the indefinite.  In the dynamic, IP-CDRT system it introduces values into the $x$ column, but for simplicity we will treat it here as merely asserting that the $x$ column is populated.) 
The $w_1$ column copies of $a_0$ are not all $\star$'s,  which is all that is required for \textbf{might} to be true. (Similar tables can be constructed with World 2, World 3, etc. in the $w*$ column.) Relative to the same table, a \textbf{must} statement would not be true, since there is a $\star$ value in the $w_1$ column, corresponding to the accessible World 4. In other words, there is no burger in World 4. (Recall that $w^*$ is the special column holding only the discourse-level world of evaluation.) 
\newpage
\ex. \a. $\textbf{might}_{w^*,a_0}^{w_1}\left([x];
	\textsc{burger}_{w_1}\{x\}\right)$
	\hfill ``there might be a burger''\label{mightburger}\\[-1ex]
	\b. \begin{tabular}{c|c|c|c}
	$w^*$ & $a_0$ & $w_1$ & $x$\\\hline
	World 1 & World 1 & World 1 & Burger 1\\
	World 1 & World 2 & World 2 & Burger 2\\
	World 1 & World 3 & World 3 & Burger 3 \\
	World 1 & World 4 & $\star$ & $\star$\\
	\end{tabular}\label{mightburger table}

Next, relative to the same table, a formula like \Next is theoretically acceptable (the 
gloss here is \textit{they are large} because the table represents multiple 
possible burgers):
\ex. $\textsc{large}_{w^*}\{x\}$
\mytabhere $\rightsquigarrow \lambda G \,.\, \forall g{\in}G \left( \textsc{large}_{g(w^*)}(g(x))\right)$  \hfill ``they are large''

And as long as each burger (Burgers 1--3) is large in World 1, the formula 
will be true. In other words, it is predicted to mean that the possible burgers 
just mentioned are large. As described above, though, this is an unwelcome 
result: it is odd to refer to individuals introduced under a modal like 
\textit{might}.

To solve this problem, \citet{stone1999reference} (and the others following him) turns to an idea from \cite{muskens1995tense}: while there is a single domain of individuals, each world determines a subset of this domain, those individuals that exist in that world. This notion is captured by the predicate \textbf{in}, which is true of a (single) individual and (single) world if that individual exists in that world. Although the various systems implement this idea differently, they all assume that a pronoun like \textit{it} or \textit{they} presupposes that its referent exists in the world of evaluation. 

Although the previous analyses do not formalize the presuppostion involved, we 
can sketch it out as in \Next, where the clause after ``$|$'' is a presupposition, using the same notation as PIP:\footnote{Brasoveanu's full definition is as follows, where $G_{u\neq\star,p\neq\star}$ is that subset of $G$ where no dummy values appear in the $u$ or $p$ columns:
\ex. $u \textbf{ in } p ~:=~ \lambda G \,.\, G_{u\neq\star,p\neq\star} \neq \varnothing ~\land~ \forall g {\in} G_{u\neq\star,p\neq\star} (g(u) \textbf{ in } g(p)) $

The exclusion of dummy values is important. The $\star$ is designed to falsify every predicate (including \textbf{in}), but pronouns can felicitously refer to column values that include $\star$. For instance, the $p$ column in a table verifying \Next will include dummy values (for the students who did not write papers). And yet, the pronoun \textit{they} can successfully refer (just) to the non-$\star$ values of $p$, the papers actually written.
\ex. Half the students wrote a$^p$ paper. They$_p$ are on my desk.

}
\ex. $\textsc{large}_{w*}\{x\}$ \hfill ``they are large''\\ $\rightsquigarrow \lambda G \,.\, \forall g{\in}G 
\left(\textsc{large}_{g(w^*)}(g(x)) \,\middle|\,{\textbf{in}(g(x),g(w*))}\right)$ 

Given a table $G$, \Last presupposes that all rows/assignments $g{\in}G$ are 
such that $g(x)$ exists in $g(w*)$; it asserts that in each such $g$, $g(x)$ 
is large in $g(w*)$. Since nothing guarantees that Burgers 1--3 all exist in 
World 1, previous scholars have assumed that this presupposition is not met in 
\ref{mightburger table}. (Presumably Burger 1 exists in World 1, in order for 
the first row to satisfy the formula, but the other burgers do not necessarily 
exist there.) And 
therefore, the sentence is not felicitous relative to information
state~\ref{mightburger table}.

The problem with this analysis comes, as mentioned above, when the individuals 
actually \textit{do} exist in the relevant scenario. For instance, consider an 
alteration of \ref{mightburger table} to fit the scenario introduced above 
involving the winner of an election with a fixed set of known candidates:
\ex. \a. $\textbf{might}_{w^*,a_0}^{w_1}\left([x]; \textsc{winner}_{w_1}\{x\}\right)$
\hfill ``there might be a winner''\label{mightwinner}
	\b.  $\textsc{woman}_{w^*}\{x\}$ \hfill ``she's a woman''\\
	$~\rightsquigarrow~ \forall g{\in}G \left(\textsc{woman}_{g(w^*)}(g(x)) \,\middle|\, {\textbf{in}(g(x),g(w^*))} \right)$\label{woman} \\[-1ex]
\b. \begin{tabular}{c|c|c|c}
$w^*$ & $a_0$ & $w_1$ & $x$\\\hline
World 1 & World 1 & World 1 & Jones\\
World 1 & World 2 & World 2 & Jones\\
World 1 & World 3 & $\star$ & $\star$\\
\end{tabular}\label{mightwinner table}

Here, the state is set up so that there is a single winner, namely Jones, 
across all worlds (Worlds 1\& 2) where there is a winner. There is also an 
accessible world, World 3, where there is no winner (yet).

Now, in the likely scenario that the epistemically possible winner(s) of an 
election exist in the real world, the presupposition in \ref{woman} will 
actually be met relative to the information state in \ref{mightwinner table}. 
According to this analysis, then, since Jones exists in World 1, the sentence 
\textit{she's a woman} should be felicitous after \textit{there might be a 
winner}, contrary to fact. And similar reasoning applies to a plural pronoun 
case such as \textit{they are women}, which would be felicitous if there 
were more than one possible winner.

In addition to this case of over-generation, admitting an infelicitous sentence, 
Brasoveanu's (but not Stone's) presupposition analysis using \textbf{in} suffers from under-generation as 
well.\footnote{Stone's system does not suffer this problem, since it introduces individuals relative to worlds one-by-one, and distributes over all worlds in the context set in each predication. So, the animals would be introduced and retrieved distributively world-by-world in this case. Note, however, that Stone's logic is singular, and does not treat summation pronouns or quantificational subordination, and thus Brasoveanu's system is a closer overall point of comparison to PIP.} Consider the following, illustrating a \textit{must}-statement:
\ex. \a. $\textbf{must}_{w_0}^{w_1}\left([x];\, \textsc{animal}_{w_1}\{x\};\,  \textsc{in-shed}_{w_1}\{x\}\right)$\\
\strut\hfill ``there must be an animal in the shed''\label{mustshed}
\b. \begin{tabular}{lc|c|c|c}
       & $w^*$   & $a_0$   & $w_1$   & $x$\\\hline
$g_1$: & World 1 & World 1 & World 1 & Animal 1\\
$g_2$: & World 1 & World 2 & World 2 & Animal 2\\
$g_3$: & World 1 & World 3 & World 3 & Animal 3\\
\end{tabular}\label{mustshed table}

(The rows/assignments are labeled for convenience.) The column under $w^*$ 
indicates that World~1 is the world of evaluation at the discourse level, and each world under 
$a_0$ is one that is epistemically accessible from World 1.  The 
column under $w_1$ represents all the worlds where there is an animal in the 
shed: note that there is a different possible animal in each of these 
worlds, as represented in the $x$ column. The modal \textbf{must} in 
\ref{mustshed} requires that there be an animal $x$ in the shed in the 
corresponding world in $w_1$. Crucially, though, it does 
\textit{not} require $x$ to be an animal in the shed in 
the real world, stored in $w^*$, only in the supposition world $w_1$.

Additionally, there is indeed no more evidence here that the various potential 
animals 
(Animal 1--3) exist in all Worlds 1--3 than there is that the burgers in 
\ref{mightburger table} exist in their Worlds 1--3. Each animal is only really 
guaranteed to exist in its corresponding world in $w_1$ (Animal 1 in 
World 1, Animal 2 in World 2, Animal 3 in World 3). Therefore, the 
presupposition of a pronoun in a subsequent sentence (such as \textit{it's 
making quite a racket}) would not be satisfied in a state like \ref{mustshed 
table}: 
\ex.  $\textsc{loud}_{w^*}\{x\}$ \hfill ``it's / they're making quite a 
racket''\\$\rightsquigarrow~ \forall g{\in}G \left(\textsc{loud}_{g(w^*)}(g(x)) \,\middle|\, {\textbf{in}(g(x),g(w^*))}\right)$

In particular, the presupposition
${\textbf{in}(g(x),g(w*))}$ would fail due to the
assignments $g_2$ and $g_3$ in \ref{mustshed table}: $g_2(x)$ and
$g_3(x)$ retrieve animals that do not exist in $g_2(w^*)$ and
$g_3(w^*)$, that is, in World 1. Therefore, this sentence is predicted to
be infelicitous, again contrary to fact. 

Due to these two counterexamples, we hold that the solution to Stone's anaphora 
problem cannot lie in merely requiring the individuals in question to exist in 
the relevant world of evaluation. Rather, the individuals must also satisfy 
\textit{in the evaluation world} the antecedent description predicated of them in the 
supposition worlds. In the examples above, 
in order to be referred to at the discourse level, the burger, winner, or animal-in-the-shed
$x$ must actually be a burger, winner, or animal-in-the-shed in world $w^*$.

\subsection{Eliott, Hofmann, Mandelkern}\label{mostrecent}
We have concentrated so far on IP-CDRT, since it is the system closest in empirical coverage to our own (PIP). \citet{brasoveanu2010decomposing} provides analyses for most of the phenomena that PIP captures: plurals, intensions, quantificational and modal subordination, and summation pronouns. IP-CDRT does not capture paycheck pronouns, which are not as simple for dynamic systems \citep{nouwen2020type}, and it does not provide an explicit representation of presuppositions like PIP, but otherwise it is quite comparable. 

In this subsection, we present some recent work
\citep{elliott2022partee,hofmann2022diss,mandelkern2022witnesses} 
on dynamic systems that preserve more information about antecedent
descriptions, and that could in principle provide an analysis of the
examples we have been considering (though none of them have yet
actually done so).%

What all three of these
approaches have in common is the ability to link the definedness of a
variable to the truth of a proposition within a single table or
information state. For instance, imagine a table with columns for
worlds $w$ and for bathrooms $b$. In these systems, the $w$ column
might contain some worlds with bathrooms and some worlds without; but
the $b$ column will be undefined when the corresponding world in $w$
does not have a bathroom. In other words, $b$ is only defined in $w$
when the proposition \textit{there is a bathroom} is true in $w$. 

The distinction we have drawn in this paper between value-based and
de\-scrip\-tion-based systems is a matter of degree, based on how much
information is preserved from the original description. The approaches
just mentioned lie much closer to the description-based end of the scale.
Although they do not preserve the original antecedent description literally
(\textit{there is a bathroom}),
they do preserve significantly more information about it. In
particular, they precompute tables of what the description evaluates to in negative and modal
contexts, forestalling the need to have the description in hand when
those contexts arise.
We therefore do not offer significant specific critiques of
these approaches: they simply offer an alternative descriptivist
account, and both accounts could be of interest to semanticists
working in the area.  

However, none of these systems are plural logics: they do not address plurals, summation or paycheck pronouns, or presuppositions (beyond felicity conditions for variables/pronouns). In addition, \cite{mandelkern2022witnesses} does not address modals at all, and \cite{hofmann2022diss} only does so in passing. \cite{elliott2022partee} does account for modals, but does not discuss other major components of modern semantics, such as generalized quantifiers. And, as explained in Section \ref{sec:related}, Stone's intensional anaphora cases pose particular problems once summation pronouns are introduced. Therefore, some nontrivial work remains before these systems can truly be compared apples-to-apples with PIP.\footnote{Even though \cite{hofmann2022diss} is a singular logic based on a plural logic, decisions obviously remain about how precisely to reintroduce plurals, while maintaining the new empirical coverage.}

Turning first to \cite{mandelkern2022witnesses}, this work follows a
similar path to PIP: there is an almost completely classical
logic,\footnote{Mandelkern's logic adds two new quantifiers, but
both can be reduced to completely classical syntax. The
interpretation of his logic also makes use of partial assignments,
but these could be converted to total assignments, as well.} which
is used to determine a sentence's truth conditions; but this logic is
paired with a separate recursive procedure to determine the felicity
of definite descriptions (including pronouns). This separate
procedure, which Mandelkern terms ``satisfying %
bounds'' or \textbf{satt} tracks a set of assignment/world pairs akin
to an information state through a discourse. Simplifying greatly, a
sentence $p$ containing an indefinite indexed $x$ shapes this
information state as follows: the assignment/world pairs in the state
where $p$ is true are guaranteed to have defined values for $x$ while
those where $p$ is false are not. Next, sentences $q$ containing a
definite description indexed $x$ (where $x$ must be free in $q$) require $x$ to satisfy $q$ throughout this entire information state. But since Mandelkern assumes a starting state containing all assignment/world pairs, including partial assignments where $x$ is undefined, this requirement will not be met unless a previous sentence includes an indefinite indexed $x$, removing undefined values. Mandelkern does briefly discuss definitions for \textbf{satt} suitable for generalized quantifiers, but does not include plurals, modals, or any phenomenon more complex than these.

Turning next to \cite{elliott2022partee}, this paper presents a solution to what Elliott terms \textbf{Partee Conjunctions}, sentences like \textit{Maybe there's no bathroom in this house, and maybe it's in a funny place.} Elliott constructs a Bilateral Update Semantics: bilateral since it presents separate functions for positive and negative truth values, and an update semantics since its interpretation functions take and return a single, complex state. These states are sets of pairs each comprising a world and an assignment function. The initial state assigns no values to variables, but each existential statement $\exists x \phi$ adds values for its variable $x$.

In Elliott's system, updating a state on an epistemic \textit{might} statement $\Diamond\phi$ will succeed (produce a non-empty output state) as long as $\phi$ is true in some subset of the input state state $s$. When it succeeds, it also updates the state to add any new variables introduced in $\phi$---but crucially \emph{only} in that portion of $s$ where $\phi$ holds. For instance, if $s$ contains no values for $x$, updating with $\Diamond(\exists x\, \textsc{winner}(x))$ will output a state $s'$ containing all pairs $\tp{w,g}$ where $g(x)$ is a winner in $w$; $s'$ will also contain the remaining world/assignment pairs in $s$ where there is no winner, too, just without values for $x$. In such a mixed state, where some assignments have values for $x$ and others do not, a statement like $\textsc{woman}(x)$ is not defined: Elliott requires the variables in a predication to be defined in each component assignment of its input state to succeed. This derives Stone's observed behavior for anaphora out of epistemic modals, since the sequence \textit{There might be a$^x$ winner; she$_x$ is a woman} does indeed come out as infelicitous/undefined for Elliott.

Although Elliott's system succeeds for this case, as mentioned above, it is not straightforwardly clear how it would extend to other, related phenomena such as non-epistemic modal statements, and modal subordination, let alone true generalized quantifiers, plurals, summation pronouns, and paycheck pronouns. Therefore, we will devote the rest of this subsection to a proposal by \citet{hofmann2019anaphoric,hofmann2022diss}, who updates the work by Stone and Brasoveanu to better capture the relationship between disjunction, negation, and anaphora; although her system does not give analyses for plurals, modals, subordination, or presupposition projection (beyond simple definedness of variables).

In particular, Hofmann gives an intensional, dynamic semantics based mainly on \cite{brasoveanu2010decomposing}, but without the plurals. She follows \cite{stone1999reference} and \cite{stone1999dynamic}, though, in two areas.
First, she makes discourse referents individual concepts (functions from worlds to individuals) instead of direct individuals. Second, she relativizes the introduction of such discourse referents to a particular proposition / set of worlds. Translating slightly to our notation, Hofmann's definition is given below (where $\star$ represents an undefined value):

\ex. Hofmann's Relative Variable Update: $g[\phi : x]h$ iff \begin{samepage}
	\a. $g[x]h$,
	\b. $\forall w(h(\phi)(w) \rightarrow h(x)(w) \neq \star))$, and
	\b. $\forall w(\lnot h(\phi)(w) \rightarrow h(x)(w) = \star))$
	\z.
\end{samepage}

The variable $\phi$ here denotes a proposition (function from worlds to truth values) and $x$ is an individual concept. Clause (a) is the usual dynamic random assignment to introduce $x$, but clauses (b-c) require $x$ to be defined whenever $\phi$ is true, and undefined whenever $\phi$ is false. The result is quite similar to Elliott's update for epistemic modals.

Hofmann's predicates, too, are defined relative to a proposition, which is maximized by various elements in the discourse. For instance, here is (a simplified version of) her analysis for a negated sentence, showing how negation maximizes the proposition associated with its prejacent (Hofmann writes $\overline{\phi}$ for the complement of $\phi$, that is, those worlds not in $\phi$):
\ex. \a. \begin{minipage}[t]{\textwidth} $\textsc{bathroom}_\phi\{x\} ~\stackrel{g}{\rightsquigarrow}~ \forall w (g(\phi)(w) \rightarrow \textsc{bathroom}_w(g(x)(w)))$ \end{minipage}
	\b. There is no$^x$ bathroom\\$~~~\rightsquigarrow  [\psi]; \phi {=} \overline{\psi}; \textbf{max}_\psi([\psi:x]; \textsc{bathroom}_{\psi}\{x\})$\label{nobath}\z.

After updating on \ref{nobath}, $\psi$ will be the maximal set of worlds where there is a bathroom, and $\phi$ is the complement of this set, that is, the worlds where there is no bathroom. The individual variable $x$ will only be defined for worlds in $\psi$ (that is, worlds where there is a bathroom). 

Hofmann also tracks a speaker's discourse commitments, and the assertion of \ref{nobath} has the effect of narrowing the set of worlds $\phi_{\textsc{DC}_\textsc{S}}$ that the speaker takes as candidates for the real world. (Assertion also maximizes a proposition, although that has little effect here.) Thus, if a future sentence tries to update speaker commitments again with worlds where the bathroom exists, we will derive a contradiction:
\ex. \a. \textsc{Assert} there is no$^x$ bathroom.\\$\rightsquigarrow [\phi]; \phi_{\textsc{DC}_\textsc{S}} {\subseteq} \phi; \textbf{max}_\phi([\psi]; \phi {=} \overline{\psi}; \\ \phantom{\rightsquigarrow\ }\textbf{max}_\psi([\psi:x]; \textsc{bathroom}_{\psi}\{x\}))$
	\b. \begin{minipage}[t]{\textwidth}\textsc{Assert} it$_x$ is huge.\\ $\rightsquigarrow [\phi']; \phi_{\textsc{DC}_\textsc{S}} {\subseteq} \phi'; \textbf{max}_{\phi'}(\textsc{huge}_{\phi'}\{x\})$\end{minipage}

Specifically, Hofmann's \textbf{max} operator throws out worlds where the prejacent formula is not defined. Therefore, the proposition $\phi'$, after maximization, will only contain worlds where $x$ is defined, that is, where there is a bathroom. Asserting \Last[b] is thus similar to asserting \textit{There is a bathroom and it is huge}. However, we have already narrowed the speaker's discourse commitments in \Last[a] to only include worlds without a bathroom, leading to contradiction.

Hofmann's analysis of Partee's bathroom sentence is shown next. Note that here the speaker is not committed to either disjunct alone, only to their union.
\ex. \textsc{Assert} there is no bathroom or it is in a funny place.\\
	$\rightsquigarrow [\phi_1]; \phi_{\textsc{DC}_\textsc{S}} {\subseteq} \phi_1;\\ \phantom{\rightsquigarrow\ } \textbf{max}_{\phi_1}\hspace{-.15cm} \left(\begin{array}{l}
		[\phi_2, \phi_3]; \phi_1{=}\phi_2{\cup}\phi_3;\\ \textbf{max}_{\phi_2}\left(\begin{array}{l}[\phi_4]; \phi_2 {=} \overline{\phi_4};\\ \textbf{max}_{\phi_4}([\phi_4\!:\!x]; \textsc{bathroom}_{\phi_4}\{x\})\end{array}\right);\\
		\textbf{max}_{\phi_3}(\textsc{funny-place}_{\phi_3}\{x\})
		\end{array}\right)$

The crucial part for Hofmann is that the clause ``$\textbf{max}_{\phi_3}(\textsc{funny-place}_{\phi_3}\{x\})$'' throws out the worlds where $x$ is not defined, again leaving just the worlds where there is a bathroom. So, $\phi_1$ ends up as the union of the maximal set of worlds without a bathroom ($\phi_2$) and the maximal set of worlds where there is a bathroom and it is in a funny place ($\phi_3$).

Although Hofmann does not present an analysis for modals, one rather straightforward treatment in her system would be to adapt (like Elliot) an idea from \cite{veltman1996defaults}: namely, that an epistemic statement ``might $\phi$'' tests whether the current state is consistent with $\phi$. For modals, Hofmann could replace her treatment of declarative mood assertions with a clause testing the consistency of the speaker's discourse commitments with a prejacent proposition, as shown in \Next:
\ex. \textsc{Might} there is a$^x$ bathroom.\\ $\rightsquigarrow [\phi]; (\phi_{\textsc{DC}_\textsc{S}} {\cap} \phi \neq \varnothing);  \textbf{max}_\phi([\phi:x]; \textsc{bathroom}_{\phi}\{x\}))$\label{mightbath}

The formula in \Last will only succeed if the speaker is committed to there being at least some worlds under discussion where there is a bathroom (some subset of $\phi$). Of course, there could also be worlds where there is not a bathroom. Either way, the value for $x$ is not changed from above: it is only defined for the worlds in $\phi$, and therefore only defined for those worlds in \(\phi_{\textsc{DC}_\textsc{S}} \) where there is a bathroom, just as in Elliott's proposal.

This is where Hofmann's system could run into problems for the cases considered here, though. Recall that her \textbf{max} operator crucially discards undefined cases, yielding the correct results for Partee's disjunction. This leads to overgeneration in the modal case, however. Consider the following assertion, which sounds odd after the modal statement in \ref{mightbath}:
\ex. \textsc{Assert} it$_x$ is in a funny place.\\
	$\rightsquigarrow [\psi]; \phi_{\textsc{DC}_\textsc{S}} {\subseteq} \psi; \textbf{max}_\psi(\textsc{funny-place}_{\psi}\{x\})$\label{funnyplace}

For Hofmann, there should be no contradiction, just like in a
disjunction case. Here, $\psi$ is the maximal set of worlds where $x$
is (defined and) in a funny place. In other words, $\psi$ is the set
of worlds where
there is a bathroom and it is in a funny place. Asserting \Last
therefore makes \(\phi_{\textsc{DC}_\textsc{S}}\) a subset of $\psi$,
throwing out any worlds in $\phi_{\textsc{DC}_\textsc{S}}$ where there
is no bathroom, just as if \ref{funnyplace} read \textit{There is a
  bathroom and it's in a funny place.} This incorrectly predicts
\ref{funnyplace} to be felicitous.\footnote{
Of course, Hofmann might
object to the analysis for \textit{might} shown above, preferring one
where the assertion in \ref{mightbath} requires some worlds where
there is a bathroom and some where there is not, leading to
contradiction in \Last. To the extent that this is a meaning of
\ref{mightbath}, it seems to be
the result of a scalar implicature that strengthens ``might'' to
``might but not must.'' Namely, cancellation (of the implicature) is
indeed possible:

\ex. There might be a bathroom here. (\textit{Looks around the corner.}) In fact, there is!

But if the ``might but not must'' reading is the result of a
cancelable scalar implicature, under Hoffmann's analysis
\ref{funnyplace} or \ref{intsum} (repeated as {\Next}) should be felicitous:

\ex. Andrea might be eating a burger. \#It is very large (in fact)!

Namely, it should be possible to
cancel the implicature that there are non-burger worlds in order to
correctly update with \textit{it is large}. Empirically, that
prediction is incorrect.

} 

As mentioned above, one feature that unites all three of these works
is that they allow definedness of a variable to depend on a particular proposition or formula. This is halfway towards a description-based account of anaphora: the variable is only defined under the description provided by the proposition in question. The successes of these accounts then support the main thesis of this paper, namely that the description-based account provides a simpler analysis to the problem at hand. However, our analysis takes this hypothesis to its natural conclusion, arguing that description-based anaphora, when paired with the bare minimum to support intensions, presuppositions, and extended scope for indefinites, can single-handedly account for all the standard puzzles in plural semantics, plus the intensional anaphora puzzles of \cite{stone1997anaphorictense}. It is ``one weird trick'' for plural semantics.\footnote{See, e.g., Alex Kaufman, \textit{Prepare to Be Shocked! What happens when you actually click on one of those ``One Weird Trick'' ads?}, \textbf{Slate: Moneybox} (July 30, 2013),  \url{https://slate.com/business/2013/07/how-one-weird-trick-conquered-the-internet-what-happens-when-you-click-on-those-omnipresent-ads.html}, archived at \url{https://perma.cc/8VNQ-T5UN} .}

\section{Conclusion}

To summarize, we have proposed an account of anaphora to antecedents
in intensional contexts that addresses problems with the proposals of
\citet{stone1999reference,stone1999dynamic}, and
\citet{brasoveanu2010decomposing}. The system that we employ, plural
intensional presuppositional predicate calculus (PIP), adds only
standard implementations of intensions, pluralities, and
presuppositions to systems previously proposed to account for complex
anaphora, including donkey anaphora, anaphora to indefinites under
quantification, and paycheck pronouns
\citep{keshet2018dynamic,unif}. These additions suffice to capture the
phenomena examined, without specially tailored
definitions. Furthermore, PIP (unlike its predecessors) is entirely
grounded in standard predicate calculus with set abstraction.

A key difference between our analysis and those of Stone, Hardt, and
Bra\-so\-ve\-anu can be characterized as a contrast between  {\it description-based\/} and {\it value-based\/} approaches to anaphora. In our account, formula labels provide the mechanism of reference to descriptions. (Other recent work also adopts proposals that incorporate descriptions for different phenomena.) Another key feature of our analysis is that, although modal and quantificational contexts both involve generalized quantifiers, the world variable that
is quantified over in a modal context is an external variable, whereas
the variable that is quantified over in quantificational contexts is a
local variable; that distinction explains the empirical differences
between anaphora to indefinites under modals versus anaphora to
indefinites under quantifiers.

%
%
%
%

\newpage
\nocite{Heim:2011}
\printbibliography

\begin{addresses}
  \begin{address}
    Ezra Keshet \\
    Department of Linguistics \\
    University of Michigan \\
    440 Lorch Hall\\
    611 Tappan Street\\
    Ann Arbor, MI 48109-1220 \\
    \email{ekeshet@umich.edu}
  \end{address}
  \begin{address}
    Steven Abney \\
    Department of Linguistics \\
    University of Michigan \\
    440 Lorch Hall\\
    611 Tappan Street\\
    Ann Arbor, MI 48109-1220 \\
    \email{abney@umich.edu}
  \end{address}
\end{addresses}

\end{document}